\documentclass[final]{cvpr}

\makeatletter
\@namedef{ver@everyshi.sty}{}
\makeatother
\usepackage{tikz}
            
\usepackage{nicefrac}       
\usepackage{microtype}      

\usepackage{enumitem}
\usepackage{times}
\usepackage{epsfig}
\usepackage{graphicx}
\usepackage{mathptmx}
\usepackage{amsmath}
\usepackage{amssymb}
\usepackage{eucal}
\usepackage{float}
\usepackage[linesnumbered, ruled, vlined]{algorithm2e}
\usepackage{subcaption}
\usepackage{tabularx}
\DeclareMathAlphabet{\mathcal}{OMS}{cmsy}{m}{n}
\usepackage{booktabs}
\usepackage{multirow}
\usepackage[pagebackref=true,breaklinks=true,colorlinks,bookmarks=false,linkcolor=blue,citecolor=blue]{hyperref}

\def\cvprPaperID{****} 
\def\confYear{CVPR 2021}
\tikzstyle{arrow} = [thick,->,>=stealth]
\tikzstyle{process} = [rectangle, minimum width=2cm, minimum height=1cm, text centered, draw=black]
\makeatletter
\tikzset{
    database/.style={
        path picture={
            \draw (0, 1.5*\database@segmentheight) circle [x radius=\database@radius,y radius=\database@aspectratio*\database@radius];
            \draw (-\database@radius, 0.5*\database@segmentheight) arc [start angle=180,end angle=360,x radius=\database@radius, y radius=\database@aspectratio*\database@radius];
            \draw (-\database@radius,-0.5*\database@segmentheight) arc [start angle=180,end angle=360,x radius=\database@radius, y radius=\database@aspectratio*\database@radius];
            \draw (-\database@radius,1.5*\database@segmentheight) -- ++(0,-3*\database@segmentheight) arc [start angle=180,end angle=360,x radius=\database@radius, y radius=\database@aspectratio*\database@radius] -- ++(0,3*\database@segmentheight);
        },
        minimum width=2*\database@radius + \pgflinewidth,
        minimum height=3*\database@segmentheight + 2*\database@aspectratio*\database@radius + \pgflinewidth,
    },
    database segment height/.store in=\database@segmentheight,
    database radius/.store in=\database@radius,
    database aspect ratio/.store in=\database@aspectratio,
    database segment height=0.1cm,
    database radius=0.25cm,
    database aspect ratio=0.35,
}
\makeatother

\setitemize{noitemsep,topsep=5pt,parsep=1pt,partopsep=1pt}

\begin{document}

\title{SelfAugment: Automatic Augmentation Policies for Self-Supervised Learning}

\author{Colorado J Reed\thanks{equal contribution; correspondence to \texttt{cjrd@cs.berkeley.edu}}$^{*\dagger}$,~~Sean Metzger$^{*\ddag}$,~~Aravind Srinivas$^{\dagger}$,~~Trevor Darrell$^{\dagger}$,~~Kurt Keutzer$^{\dagger}$\\
{\footnotesize $^{\dagger}$BAIR, Department of Computer Science, UC Berkeley} \\
{\footnotesize $^{\ddag}$Graduate Group in Bioengineering (Berkeley/UCSF), Weill Neurosciences Institute \& UCSF Neurological Surgery}\\
}

\maketitle

\begin{abstract}
A common practice in unsupervised representation learning is to use labeled data to evaluate the quality of the learned representations.
This supervised evaluation is then used to guide critical aspects of the training process such as selecting the data augmentation policy.
However, guiding an unsupervised training process through supervised evaluations is not possible for real-world data that does not actually contain labels (which may be the case, for example, in privacy sensitive fields such as medical imaging).
Therefore, in this work we show that evaluating the learned representations with a self-supervised image rotation task is highly correlated with a standard set of supervised evaluations (rank correlation $> 0.94$).
We establish this correlation across hundreds of augmentation policies, training settings, and network architectures and provide an algorithm (SelfAugment) to automatically and efficiently select augmentation policies without using supervised evaluations.
Despite not using any labeled data, the learned augmentation policies perform comparably with augmentation policies that were determined using exhaustive supervised evaluations. 
\end{abstract}

\section{Introduction}
Self-supervised learning, a type of unsupervised learning that creates target objectives without human annotation, has led to a dramatic increase in the ability to capture salient feature representations from unlabeled visual data. 
So much so, that in an increasing number of cases these representations outperform representations learned from the same data with labels~\cite{chen_simple_2020, henaff2019data, he2019momentum}. At the center of these advances is a form of \emph{instance contrastive learning} where a single image is augmented using two separate data augmentations, and then a network is trained to distinguish which augmented images originated from the same image when contrasted with other randomly sampled augmented images, see~\cite{chen_simple_2020, he2019momentum, chen_improved_2020, wu2018unsupervised}.

\begin{figure}[t]
\begin{center}
    \includegraphics[width=\linewidth]{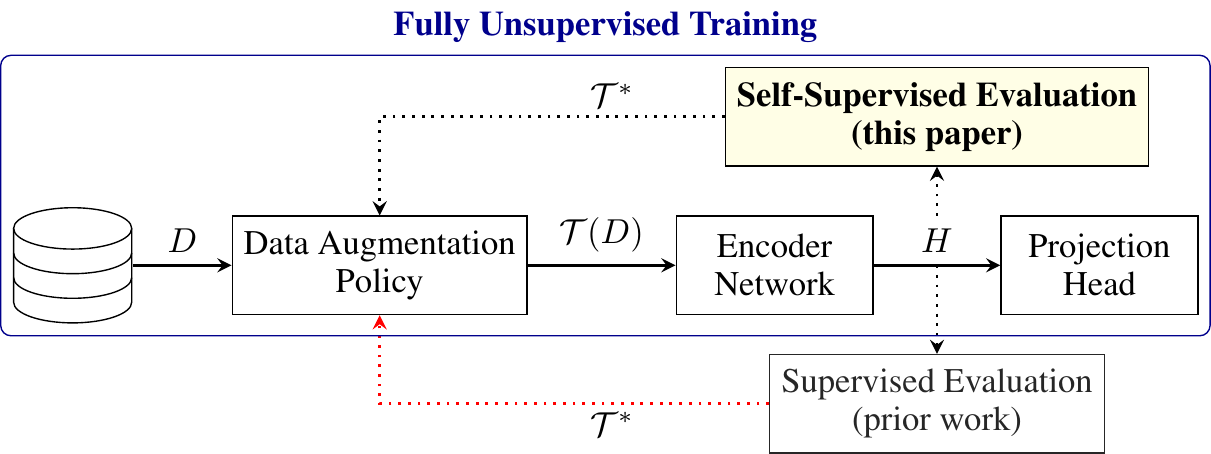}
\end{center}
   \caption{The blue box highlights our fully unsupervised training pipeline for instance contrastive representation learning: data $D$ are augmented with policy $\mathcal{T}$, then encoded into representations, $H$, which are fed into a projection head yielding features that determine the InfoNCE loss (Eq.~\ref{eq:infonce_loss}). As shown by the red arrow, prior work uses supervised evaluations of the representations to inform the training process, e.g.~to update the augmentation policy $\mathcal{T}\rightarrow\mathcal{T}^*$. In this paper, we show that self-supervised evaluation can be used in lieu of supervised evaluation and show how to use this evaluation for automatic and efficient augmentation selection.}
\label{fig:aspipe}
\end{figure}

As illustrated in Figure~\ref{fig:aspipe}, recent works~\cite{chen_simple_2020, chen_improved_2020, tian2020makes} have used extensive \emph{supervised} evaluations to determine which augmentation policies to use for training. The best policies obtain a \emph{sweet spot}, where the augmentations make it difficult for the contrastive task to determine the corresponding image pairs while retaining salient image features; finding this sweet spot can be the difference between state-of-the-art performance or poor performance for various tasks \cite{tian2020makes}.

However, it is often difficult or impossible to obtain accurately labeled data in privacy sensitive fields (e.g.~medical imaging~\cite{medical}), applications with highly ambiguous label definitions (e.g.~fashion or retail categorization~\cite{zhang2017imaterialist}), or not practical when one set of representations is used for a diverse set of downstream tasks (e.g. in autonomous driving systems~\cite{yang2018end}).
This leads to the open question: \emph{How can we evaluate self-supervised models, especially to efficiently select augmentation policies, when labeled data is not available?} We address this question via the following contributions:

\begin{itemize}[leftmargin=*]
    \item We show that a linear, image-rotation-prediction evaluation task is highly correlated with the downstream supervised performance (rank correlation $\rho > 0.94$) on six standard recognition datasets 
    (CIFAR-10~\cite{c10}, SVHN~\cite{goodfellow2013multi}, ImageNet~\cite{ILSVRC15}, PASCAL~\cite{pascal-voc-2007}, COCO~\cite{lin2014microsoft}, Places-205~\cite{zhou2014learning}) 
    and tasks (image classification, object detection, and few-shot variants)
    across hundreds of learned representations, spanning three types of common evaluation techniques: linear separability performance, semi-supervised performance, and transfer learning performance. 
    
    \item Using self-supervised evaluation, we adapt two automatic data augmentation algorithms for instance contrastive learning. Without using labeled evaluations, these algorithms discover augmentation policies that match or outperform policies obtained using supervised feedback and only use a fraction of the compute.
    
    \item We further show that using linear image rotation prediction to evaluate the representations works across network architectures, and that image rotation prediction has a stronger correlation with supervised performance than a jigsaw~\cite{noroozi2016unsupervised} or color prediction~\cite{zhang2016colorful} evaluation task.
\end{itemize}
Based on these contributions and experiments, we conclude that image rotation prediction is a strong, unsupervised evaluation criteria for evaluating and selecting data augmentations for instance contrastive learning.

\section{Background and Related Work}
In this paper, we study evaluations for self-supervised representations, particularly through the lens of learning data augmentation policies. We discuss these topics next.

\textbf{Self-supervised representation learning:}
The general goal of representation learning is to pre-train a network and then either fine-tune it for a particular task or transfer it to a related model, e.g.~see~\cite{henaff2019data, he2019momentum, reed2021self,  donahue2014decaf, erhan2010does, zeiler2014visualizing, goodfellow2016deep, lecun2015deep, radford2018improving, devlin2018bert}. Recently, \cite{chen_simple_2020} and \cite{chen_improved_2020} demonstrated substantial improvements by using similar forms of instance contrastive learning whereby one image is augmented using two separate data augmentations and then a network is trained to identify this positive pair contrasted with a large set of distractor images. A common loss function for contrastive methods is the InfoNCE loss, where given two images originating from the same image $i$ and $K_d$ distractor images we have:
\begin{equation}\label{eq:infonce_loss}\small
     \mathcal{L}_{NCE} = - \mathbb{E}\left[\log \frac{\text{exp}({\textit{sim}(\mathbf{z_{1,i}}, \mathbf{z_{2,i}})}}{\sum_{j=1}^{K_d}\text{exp}{(\textit{sim}(\mathbf{z_{1,i}}, \mathbf{z_{2,j}}))}}\right]
\end{equation}
where $\mathbf{z_{1,i}},\mathbf{z_{2,i}}$ are the two different image representations from image $i$ following the encoder network and projection head as shown in Figure~\ref{fig:aspipe}, and $\textit{sim}(\cdot,\cdot)$ is a similarity function such as a weighted dot product.

The InfoNCE loss \cite{oord2018representation, henaff2019data} has been shown to maximize a lower bound on the mutual information $I(\mathbf{h_1}; \mathbf{h_2})$. The SimCLR framework \cite{chen_simple_2020} relies on large batch sizes to contrast the image pairs, while the MoCo framework \cite{chen_improved_2020} maintains a large queue of contrasting images. Given the increased adoption, broad application, and strong performance of instance contrastive learning \cite{henaff2019data, chen_simple_2020, chen_improved_2020}, we focus our work on this type of self-supervised learning, specifically using the MoCo algorithm and training procedure for experimentation~\cite{chen_improved_2020}.

\textbf{Self-supervised model evaluation} is typically done via:
\begin{itemize}[]
    \item \emph{separability}: the network is frozen and the training data trains a supervised linear model (the justification is that good representations will reveal linear separability in the data)~ \cite{coates2012learning, rotation, doersch2015unsupervised, noroozi2016unsupervised, gidaris2018unsupervised, kolesnikov2019revisiting}
    \item \emph{transferability}: the network is either frozen or jointly fine-tuned, with a transfer task model that is fine-tuned using a different, labeled dataset (the justification is that good representations will generalize to a number of downstream tasks) \cite{henaff2019data, he2019momentum, radford2018improving,  devlin2018bert}
    \item \emph{semi-supervised}, in which the network is either frozen or jointly fine-tuned using a fraction of labeled data with either the \emph{separability} or \emph{transferability} tasks mentioned above (the justification is that a small set of labeled data will benefit good representations)~\cite{henaff2019data, chen_simple_2020}.
\end{itemize}

While these evaluations characterize the unsupervised model in several ways, they have limited use for making training decisions because: \textbf{(i)} accessing labels as a part of the training process is not possible for unlabeled datasets, and \textbf{(ii)} evaluating the model on a different, labeled dataset requires an integrated understanding of the relationship between the transfer task, datasets, and models (see ~\cite{yang2018end,puigcerver2020scalable,renggli2020model}). We seek a label-free, task-agnostic evaluation.

\textbf{Learning data augmentation policies:}
Data augmentation has played a fundamental role in visual learning, and indeed, has a large body of research supporting its use~\cite{shorten_survey_2019}. In this work, we use a self-supervised evaluation to automatically learn an augmentation policy for instance contrastive models. To formulate our automatic data augmentation framework, we draw on several, equally competitive recent works in the supervised learning space \cite{cubuk_autoaugment_2019, ho_population_2019, lim_fast_2019, cubuk_randaugment_2019}, where \cite{cubuk_autoaugment_2019, ho_population_2019, lim_fast_2019} use a separate search phase to determine the augmentation policy and \cite{cubuk_randaugment_2019} use a simplified augmentation space and sample from it via a grid search. For instance contrastive learning, we adapt a search-based automatic augmentation framework, Fast AutoAugment (FAA) \cite{lim_fast_2019}, and a sampling-based approach, RandAugment \cite{cubuk_randaugment_2019}. 
We discuss these two algorithms in greater detail in the next section.

\section{Self-Supervised Evaluation and Data Augmentation}
\label{methods}

Our central goals are to \textbf{(i)} establish a strong correlation between a self-supervised  evaluation task and a supervised evaluation task commonly used to evaluate self-supervised models, and \textbf{(ii)} develop a practical algorithm for self-supervised data augmentation selection. The following subsections defines these goals in more detail.

\subsection{Self-supervised evaluation}
\label{s:ss-eval-corr}
With labeled data, augmentation policy selection can directly optimize the supervised task performance \cite{ho_population_2019, lim_fast_2019}. 
With unlabeled data, we seek an evaluation criteria that is highly correlated with the supervised performance without requiring labels. 
Inspired by \cite{liu2020labels}, where the authors show that self-supervised tasks can be used to evaluate network architectures, we investigate the following self-supervised tasks to evaluate representations:

\begin{itemize}[leftmargin=*]
    \item \textbf{rotation} \cite{rotation}: the input image undergoes one of four preset rotations $\{0^o, 90^o, 180^o, 270^o\}$, and the evaluation metric is the 4-way rotation prediction classification accuracy
    \item \textbf{jigsaw} \cite{noroozi2016unsupervised}: the four quadrants of the input image are randomly shuffled into one of $4!=24$ permutations, and the evaluation metric is the 24-way classification accuracy
    \item \textbf{colorization} \cite{zhang2016colorful}: the input is a grayscale image, and the evaluation metric is formulated as a pixel-wise classification on pre-defined color classes (313, from \cite{zhang2016colorful})
\end{itemize}

A key point to emphasize is that these self-supervised tasks are used to \emph{evaluate} the representations learned from instance contrastive algorithms, e.g.~MoCo. These self-supervised tasks were originally used to learn representations themselves, but in this work, we evaluate the representations using these tasks. In $\S\ref{experiments}$, we compute the correlation of each of these evaluations with a supervised, top-1 linear evaluation on a frozen backbone trained using a cross entropy loss on the training data.

\subsection{Self supervised data augmentation policies}
\label{ss:self-supervised-data-aug}

\begin{algorithm}[tb]
\DontPrintSemicolon
\SetKwInOut{Input}{Input}
\Input{$\left(D_{\text{train}}, K, T, B, P, \mathcal{L} \right)$}
Split $D_{\text{train}}$ into $K$-folds: $D_{\text{train}}^{(k)}=\{(D_{\mathcal{M}}^{(k)}, D_{\mathcal{A}}^{(k)})\}$ \\

\For{$a \in \mathbb{O}$}{
    Train $\theta_{moco}$ on single aug policy $\mathcal{T}_a(D^{(1)}_{\mathcal{M}})$\\
    Train $\phi_{ss}$ on $D^{(1)}_{\mathcal{M}}$ on top of frozen $\theta_{moco}$ \\
   $a^* \leftarrow \text{argmin}_{\hat{a} \in \{a, a^*\}} \mathcal{L}(\theta_{moco}, \phi_{ss} | \mathcal{T}_{\hat{a}}(D^{(1)}_{\mathcal{M}}))$ 
}

\For{$k \in \{1,\ldots ,K\}$}{
    $\mathcal{T}^{*(k)} \leftarrow \emptyset$, $(D_{\mathcal{M}},D_{\mathcal{A}}) \leftarrow (D_{\mathcal{M}}^{(k)}, D_{\mathcal{A}}^{(k)})$\\ 
    Train $\theta_{moco}$ on base aug policy $\mathcal{T}_{a^*}(D_{\mathcal{M}})$ \\
    Train $\phi_{ss}$ on $D_{\mathcal{M}}$ on top of frozen $\theta_{moco}$ \\
    \For{$t \in \{0,\ldots,T-1\}$}
    {
        $\mathcal{B} \leftarrow$ BayesOpt$(\mathcal{T},\mathcal{L}(\theta_{moco}, \phi_{ss}|\mathcal{T}(D_{\mathcal{A}})),B)$\\
        $\mathcal{T}_{t}\leftarrow $ Select top-$P$ policies in $\mathcal{B}$\\
        Merge policies via $\mathcal{T}^{*(k)} \leftarrow \mathcal{T}^{*(k)} \cup \mathcal{T}_{t}$ 
    } 
}
\Return $\mathcal{T}^{*} = \bigcup_{k}\mathcal{T}^{*(k)}$
\caption{\textbf{SelfAugment} takes as input a dataset, $D_{\text{train}}$, parameters for policy optimization (see Appendix \ref{a:notation} for definitions), and loss function $\mathcal{L}$ and returns an augmentation policy.}
\label{alg:selfaugment}
\end{algorithm}

   We study and adapt two approaches for augmentation policy selection from the supervised domain: a sampling-based strategy, RandAugment \cite{cubuk_autoaugment_2019} and a search-based strategy, Fast AutoAugment (FAA) \cite{lim_fast_2019}. Using the notation from \cite{lim_fast_2019}, let $\mathbb{O}$ represent the set of image transformations operations $\mathcal{O}:\mathcal{X}\to\mathcal{X}$ on input image $\mathcal{X}$. Following \cite{lim_fast_2019, cubuk_randaugment_2019}, we define $\mathbb{O}$ as the set: $\{$\texttt{cutout}, \texttt{autoContrast}, \texttt{equalize}, \texttt{rotate}, \texttt{solarize}, \texttt{color}, \texttt{posterize}, \texttt{contrast}, \texttt{brightness}, \texttt{sharpnes}, \texttt{shear-x}, \texttt{shear-y}, \texttt{translate-x}, \texttt{translate-y}, \texttt{invert}$\}$ (see Appendix~\ref{a:transforms} for more  details).

Each transformation $\mathcal{O}$ has two parameters: \textbf{(i)} the magnitude $\lambda$ that determines the strength of the transformation and \textbf{(ii)} the probability of applying the transformation, $p$. Let $\mathcal{S}$ represent the set of  augmentation sub-policies, where a sub-policy $\tau\in\mathcal{S}$ is defined as the sequential application of $N_{\tau}$ consecutive transformation {\small$\{\bar{\mathcal{O}}_{n}^{(\tau)}(x; p_{n}^{(\tau)},\lambda_{n}^{(\tau)}):n=1,\ldots,N_{\tau}\}$} where each operation is applied to an input 
image sequentially with probability $p$. Applying sub-policy $\tau(x)$ is then a composition of transformation $\tilde{x}_{(n)}=\bar{\mathcal{O}}_{n}^{(\tau)}(\tilde{x}_{(n-1)})$ for $n=1,\ldots,N_{\tau}$, where the full sub-policy application has shorthand $\tilde{x}_{(N_{\tau})}=\tau(x)$ and the first application is $\tilde{x}_{(0)}=x$. The full policy, $\mathcal{T}$, is a collection of $N_{\mathcal{T}}$ sub-policies, and $\mathcal{T}(D)$
represents the set of images from $D$ obtained by applying $\mathcal{T}$.

\textbf{SelfRandAugment:} RandAugment makes the following simplifying assumptions: \textbf{(i)} all transformations share a single, discrete magnitude, $\lambda \in [1,30]$ \textbf{(ii)} all sub-policies apply the same number of transformations, $N_\tau$ \textbf{(iii)} all transformations are applied with uniform probability, $p=K_T^{-1}$ for the $K_T=|\mathbb{O}|$ transformations. RandAugment selects the best result from a grid search over $(N_\tau, \lambda)$. To adapt this algorithm for instance contrastive learning, we simply evaluate the searched $(N_\tau, \lambda)$ states using a self-supervised evaluation from $\S\ref{s:ss-eval-corr}$ and refer to this as \emph{SelfRandAugment}.

\textbf{SelfAugment:} We adapt the search-based FAA algorithm to the self-supervised setting; we call this adaptation \emph{SelfAugment}. Formally, let $\mathcal{D}$ be a distribution on the data $\mathcal{X}$. For model $\mathcal{M}(\cdot|\theta):\mathcal{X}$ with parameters $\theta$, define the supervised loss as $\mathcal{L}(\theta|D)$ on model $\mathcal{M}(\cdot|\theta)$ with data $D\sim\mathcal{D}$. For any given pair of $D_{\text{train}}$ and $D_{\text{valid}}$, FAA selects augmentation policies that approximately align the density of $D_{\text{train}}$ with the density of the augmented $\mathcal{T}(D_{\text{valid}})$. This means that the transformations should help the model bolster meaningful features and become invariant to unimportant features after retraining with the augmented dataset. In practice, FAA splits $D_{\text{train}}$ into $D_{\mathcal{M}}$ and $D_{\mathcal{A}}$, where $D_{\mathcal{M}}$ is used to train the model and $D_{\mathcal{A}}$ is used to determine the policy via:
\begin{align}
    \label{eq:objective}
    \mathcal{T}^{*} = \underset{\mathcal{T}}{\text{argmin }}\mathcal{L}(\theta_{\mathcal{M}}|\mathcal{T}(D_{\mathcal{A}}))
\end{align}
where $\theta_{\mathcal{M}}$ is trained using $D_{\mathcal{M}}$. This approximates minimizing the distance between the density of $D_{\mathcal{M}}$ and $\mathcal{T}(D_{\mathcal{A}})$ by using augmentations to improve predictions with shared model parameters, $\theta_{\mathcal{M}}$; see \cite{lim_fast_2019} for derivations. FAA obtains the final policy, $\mathcal{T}^{*}$, by exploring $B$ candidate policies $\mathcal{B}=\{\mathcal{T}_{1}, \ldots, \mathcal{T}_{B}\}$ with a Bayesian optimization method that samples a sequence of sub-policies from $\mathcal{S}$ and adjusts the probabilities $\{p_{1},\ldots,p_{N_{\mathcal{T}}}\}$ and magnitudes $\{\lambda_{1},\ldots,\lambda_{N_{\mathcal{T}}}\}$ to minimize $\mathcal{L}(\theta|\cdot)$ on $\mathcal{T}(D_{\mathcal{A}})$ (see Appendix \ref{a:bayesopt} for details). The top $P$ policies from each data split are merged into $\mathcal{T}^{*}$. The network is then retrained using this policy on all training data, $\mathcal{T}^{*}(D_{\text{train}})$, to obtain the final network parameters $\theta^*$. SelfAugment has three main differences from Fast AutoAugment that we discuss next.

\textbf{Select the base policy:}  A \emph{base augmentation policy} is required to perform the first pass of training to determine $\theta_\mathcal{M}$. SelfAugment determines this policy by training a MoCo network \cite{chen_improved_2020}
for a short period of time on each of the individual transformations in $\mathbb{O}$ as well as \texttt{random-resize-crop} (the top performing single transformation in \cite{chen_simple_2020}). Each transformation is applied at every iteration, with $p=1$ and a magnitude parameter $\lambda$ stochastically set at each iteration to be within the ranges from \cite{lim_fast_2019}. Each network is trained until the loss curves separate -- we found this to be around $10\%$ of the total training epochs typically used for pre-training. The backbone is then frozen and a linear self-supervised evaluation task, $\phi_{ss}$, is trained for each network and evaluated using held out training data. The base policy is then the transformation with the best evaluation.

\textbf{Search augmentation policies:} Given the base augmentations, we split the training data into $k\text{-folds}$. For each fold, we train a MoCo network, $\theta_{moco}$, using the base augmentation, freeze the network, and train a self-supervised evaluation layer, $\phi_\text{ss}$.
We use the same Bayesian optimization search strategy as FAA to determine the policies. However, as the loss function $\mathcal{L}(\theta|D)$ cannot use supervised accuracy as in FAA, we explore four variants of a self-supervised loss function.
Appendix~\ref{a:lossfun} discusses and compares each of these loss functions in more detail and $\S\ref{discussion}$ compares these loss functions with a supervised variant:
\begin{itemize}[leftmargin=*]
    \item \textbf{Min. eval error}: $\mathcal{T}^{SS} =  {\text{argmin}_\mathcal{T}}\mathcal{L}_{\text{ss}}(\theta_{\mathcal{M}},\phi_{\text{ss}}|\mathcal{T}(D_{\mathcal{A}}))$ where $\mathcal{L}_{\text{ss}}$ is the self-supervised evaluation loss, which yields policies that would directly result in improved performance on the evaluation task if the linear layer was retrained on top of the same base network. Minimizing the self-supervised error encourages augmentation policies that bolster distinguishable image features.
    \item \textbf{Min.~InfoNCE}: $\mathcal{T}^{\text{I-min}} =  {\text{argmin}_\mathcal{T}}\mathcal{L}_{\text{NCE}}(\theta_{\mathcal{M}}|\mathcal{T}(D_{\mathcal{A}}))$ where $\mathcal{L}_{\text{NCE}}$ is the InfoNCE loss from Eq.~\ref{eq:infonce_loss}, which yields policies that make it easier to distinguish image pairs in the contrastive feature space. It is worth noting that a trivial way to distinguish image pairs is to use weak augmentations so paired images have high similarity. 
    \item \textbf{Max InfoNCE}: $\mathcal{T}^{\text{I-max}} =  {\text{argmin}_\mathcal{T}}-\mathcal{L}_{\text{NCE}}(\theta_{\mathcal{M}}|\mathcal{T}(D_{\mathcal{A}}))$ negates the previous loss function, yielding policies that make it difficult to distinguish image pairs in the feature space. Optimizing this loss function encourages a challenging augmentation policy, which may be overly challenging for training a network to learn meaningful representations.
    \item \textbf{Min} $\mathcal{L}_{\text{ss}}$ \textbf{max} $\mathcal{L_{\text{NCE}}}$: $\mathcal{T}^{\text{minmax}} =  {\text{argmin}_\mathcal{T}}\mathcal{L}_{\text{ss}} - \mathcal{L}_{\text{NCE}}$ yields policies that simultaneously maximize InfoNCE, encouraging challenging augmentation policies, and minimize $\mathcal{L}_{\text{ss}}$, encouraging distinguishable image features.
    \end{itemize}

\begin{figure*}[htb]
\centering
\begin{subfigure}{0.25\textwidth}
\centering
\hspace{1em}\textbf{CIFAR-10}
\end{subfigure}%
\begin{subfigure}{.25\textwidth}
\centering
\hspace{1em}\textbf{SVHN}
\end{subfigure}
\begin{subfigure}{.25\textwidth} 
\centering
\hspace{1em}\textbf{ImageNet}
\end{subfigure}

\begin{subfigure}{0.25\textwidth}
    \centering
    \includegraphics[width=\textwidth]{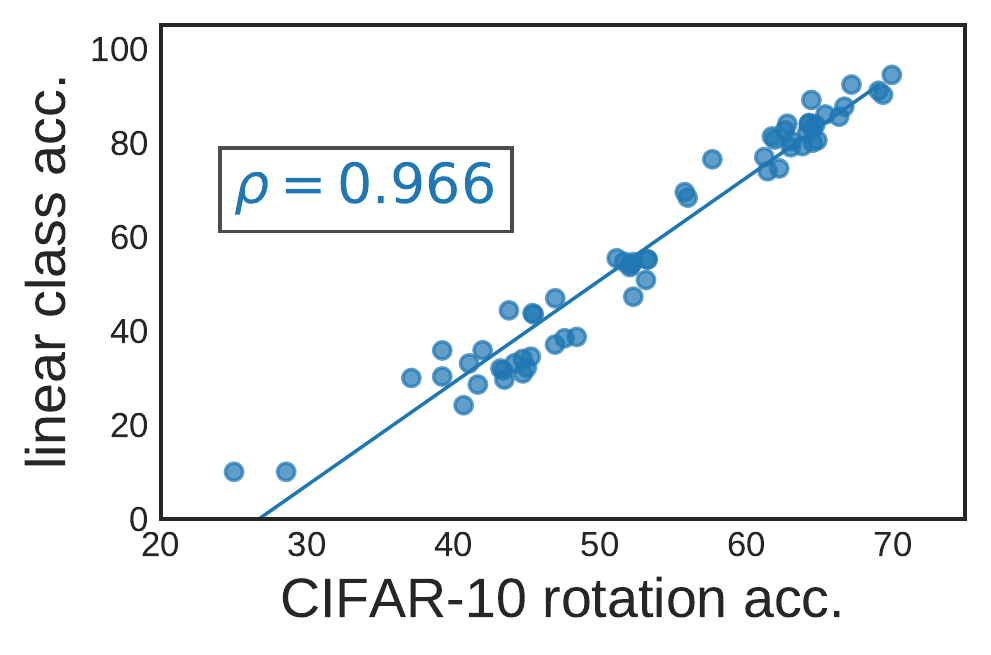}
\end{subfigure}%
\begin{subfigure}{.25\textwidth}
    \centering
    \includegraphics[width=\textwidth]{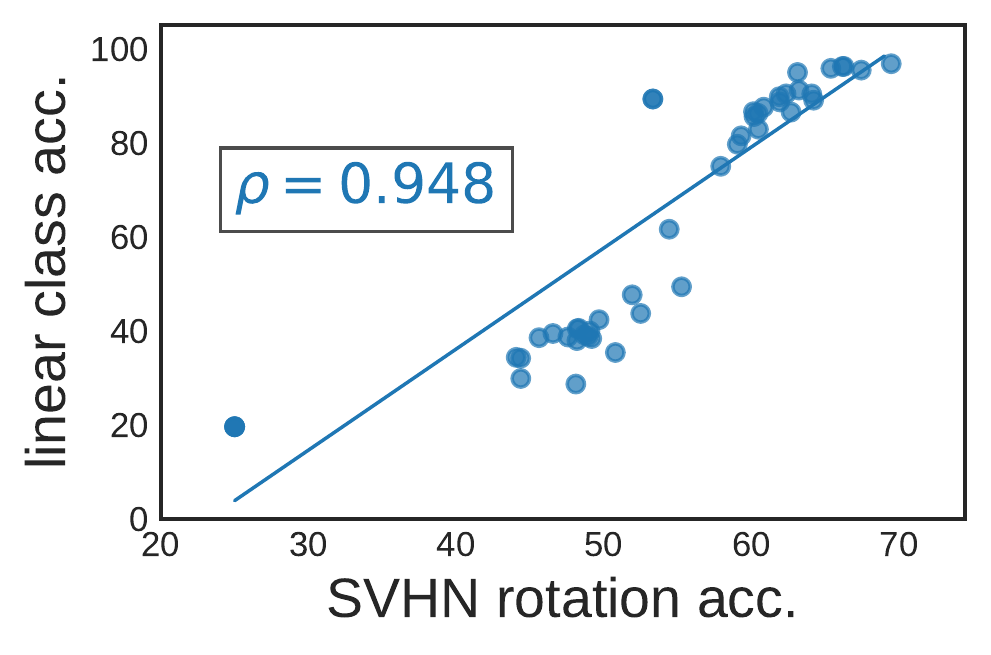}
\end{subfigure}
\begin{subfigure}{.25\textwidth}
    \centering
    \includegraphics[width=\textwidth]{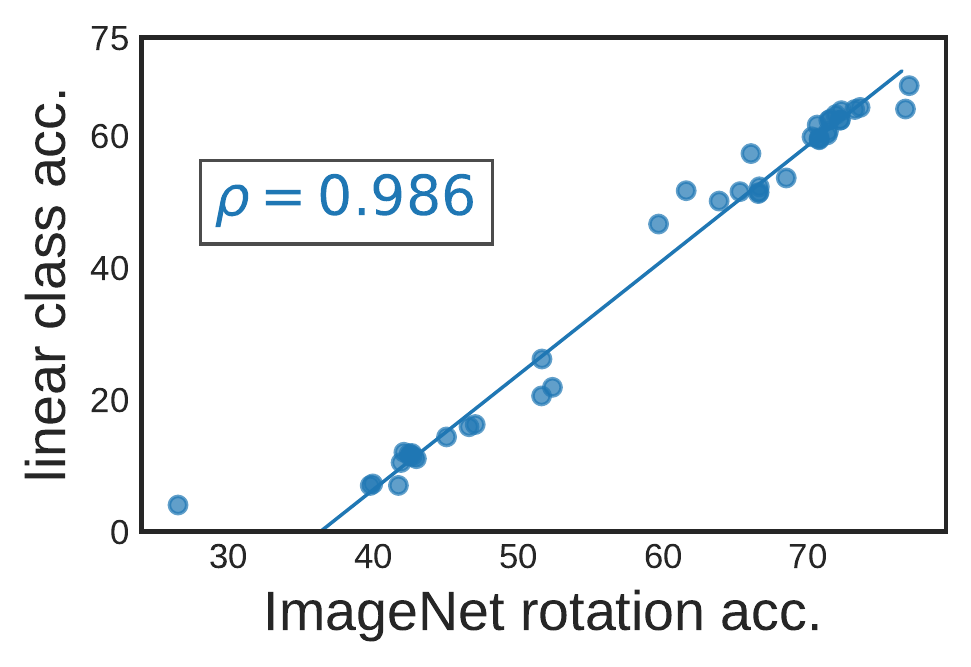} 
\end{subfigure}

\begin{subfigure}{0.25\textwidth} 
\centering
\hspace{1em}\textbf{VOC `07}
\end{subfigure}%
\begin{subfigure}{.25\textwidth} 
\centering
\hspace{1em}\textbf{CoCo 2014 }
\end{subfigure}
\begin{subfigure}{.25\textwidth}
\centering
\hspace{1em}\textbf{Places205}
\end{subfigure}

\begin{subfigure}{0.25\textwidth} 
    \centering
    \includegraphics[width=\textwidth]{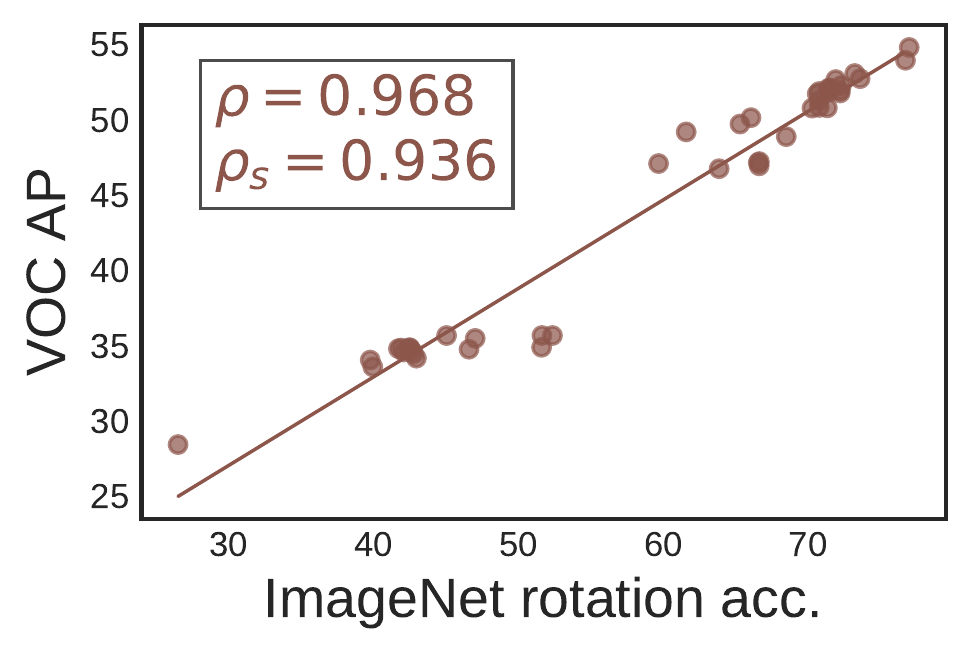}
\end{subfigure}%
\begin{subfigure}{.25\textwidth}
    \centering
    \includegraphics[width=\textwidth]{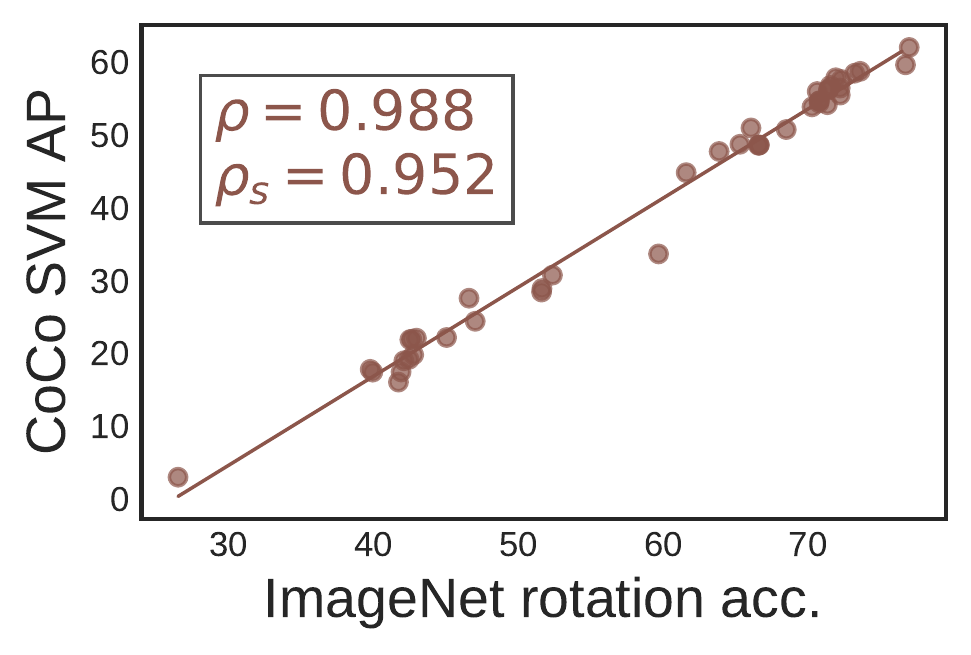}
\end{subfigure}
\begin{subfigure}{.25\textwidth}
    \centering 
    \includegraphics[width=\textwidth]{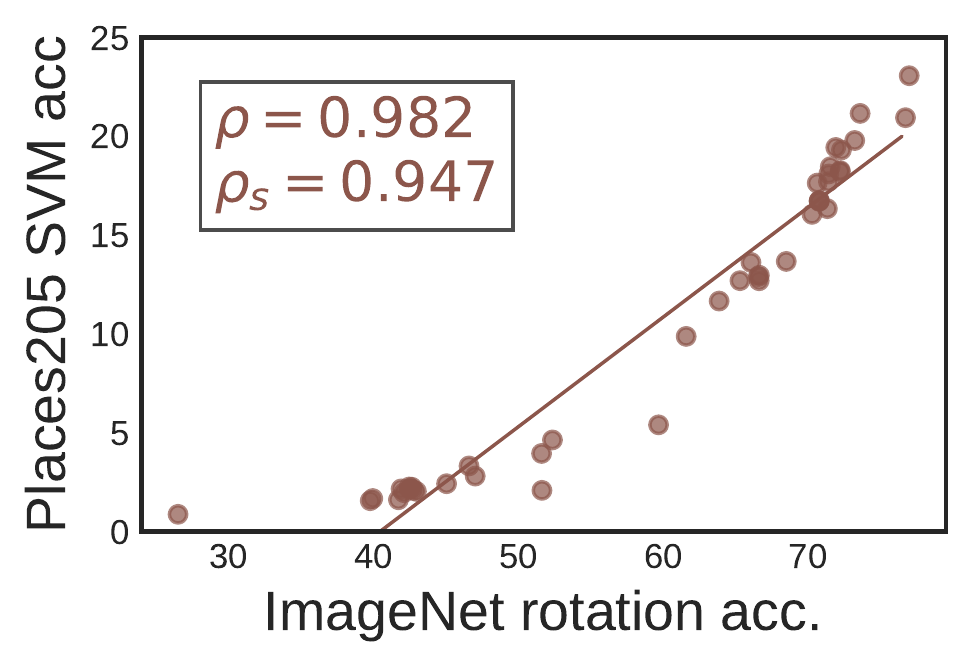}
\end{subfigure}

\caption[short]{Top row: Correlation between supervised linear classification accuracy and linear image rotation prediction accuracy for three datasets. Bottom row: Correlation between rotation prediction and three transfer tasks from the ImageNet pretraining. $\rho$ is the rank correlation between rotation prediction and the supervised task. $\rho_s$ is the rank correlation between ImageNet supervised linear classification and the transfer task performance.} 
\label{fig:corrstudy}
\end{figure*}

\textbf{Retrain MoCo using the full training dataset and augmentation policy:} SelfAugment uses the selected policy from the loss functions and then retrains from scratch on the full dataset $\mathcal{D}_{train}$. It is worth noting that because the augmentation policy learned from SelfAugment is used for an instance contrastive task, and not for the evaluation task, the augmentations that minimize the self-supervised evaluation loss are not necessarily the best augmentations for instance contrastive pre-training. Rather, this method provides the set of augmentations that would lead to high evaluation performance if the linear layer were directly retrained on top of the backbone used during augmentation selection~\cite{lim_fast_2019}. Hence, incorporating the InfoNCE loss balances the instance contrastive task with the downstream task; we observe strong empirical evidence for this in $\S\ref{experiments}$ and Appendix $\ref{a:lossfun}$.

\section{Experiments}
\label{experiments}

Through the following experiments, we aim to establish that \textbf{(i)} a self-supervised evaluation task is highly correlated with the supervised performance of standard visual recognition tasks (image classification, object detection, and few-shot variants) on common datasets (CIFAR-10~\cite{c10}, SVHN~\cite{goodfellow2013multi}, ImageNet~\cite{ILSVRC15}, PASCAL~\cite{pascal-voc-2007}, COCO~\cite{lin2014microsoft}, Places-205~\cite{zhou2014learning}) and \textbf{(ii)} SelfAugment provides a competitive approach to augmentation selection, despite being fully unsupervised. All experiments used MoCo training~\cite{chen_improved_2020} with the standard ResNet-50 backbone~\cite{szegedy2017inception} on 4 GPUs, using default training parameters from \cite{he2019momentum}, see Appendix~\ref{a:expdetails}.

\subsection{Self-supervised evaluation correlation}

As evaluating all possible data augmentations for every dataset, training schedule, and downstream task is intractable, we evaluate a diverse sampling of RandAugment, SelfAugment, MoCoV2, and single-transform policies and training schedules. For CIFAR-10~\cite{c10}, SVHN~\cite{goodfellow2013multi}, and ImageNet~\cite{ILSVRC15} we use augmentation policies: \textbf{(i)} random horizontal flip and random resize crop, \textbf{(ii)} RandAugment on top of (i) grid searched over parameters $\lambda=\{4, 5, 7, 9, 11\}, N_\tau=\{1, 2, 3\}$ at $\{100, 500\}$ epochs for CIFAR-10/SVHN and $\lambda=\{5, 7, 9, 11, 13\}$, $N_\tau=2$  at $\{20,60,100\}$ epochs for ImageNet, \textbf{(iii)} each individual RandAugment transformations at $100$ and $10$ epochs for CIFAR-10 and ImageNet, respectively, \textbf{(iv)} scaling the magnitude $\lambda$ from its min to max value for each $N_\tau$ at $500$ epochs for CIFAR-10/SVHN and $\{20,60,100\}$ epochs for ImageNet, \textbf{(v)}  for CIFAR-10/SVHN we also performed RandAugment using $K_T=\{3,6,9\}$ transformations at each of $\lambda=\{4, 7\}$ with $N_\tau=2$, at $500$ epochs, \textbf{(vi)} MoCoV2 augmentations at ${100,\;200}$ epochs for ImageNet, \textbf{(vii)} the five SelfAugment policies at $750$ epochs for CIFAR-10/SVHN and $100$ epochs for ImageNet. In total, this yields a diverse set of $61$ models for CIFAR-10/SVHN and $43$ models for ImageNet.

\begin{figure*}[t]
\centering
\begin{subfigure}{0.28\textwidth}
    \centering 
    \includegraphics[scale=0.48]{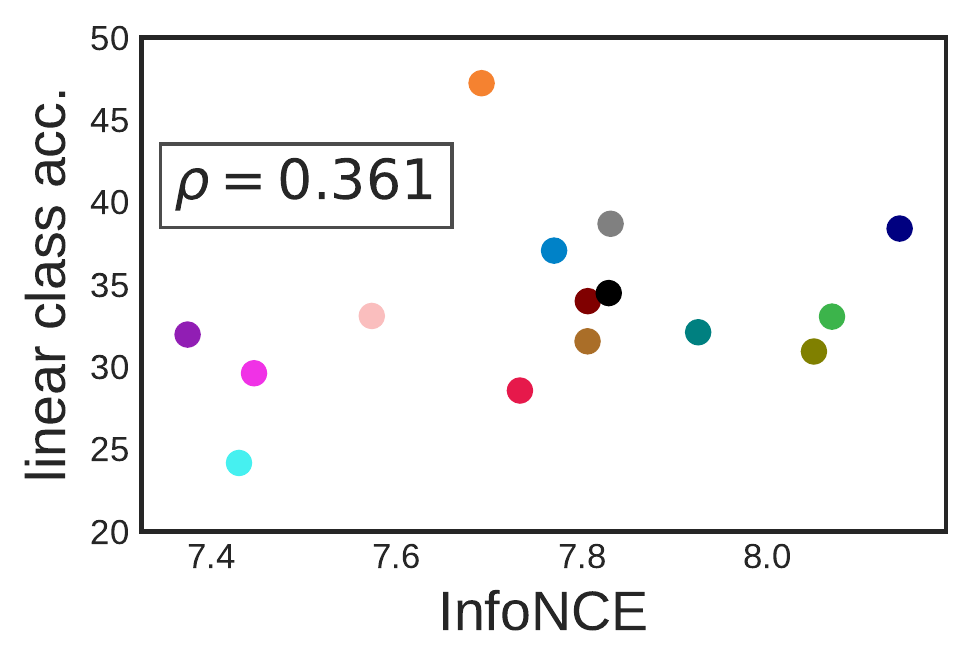}
\end{subfigure}%
\begin{subfigure}{.28\textwidth}
    \centering
    \includegraphics[scale=0.48]{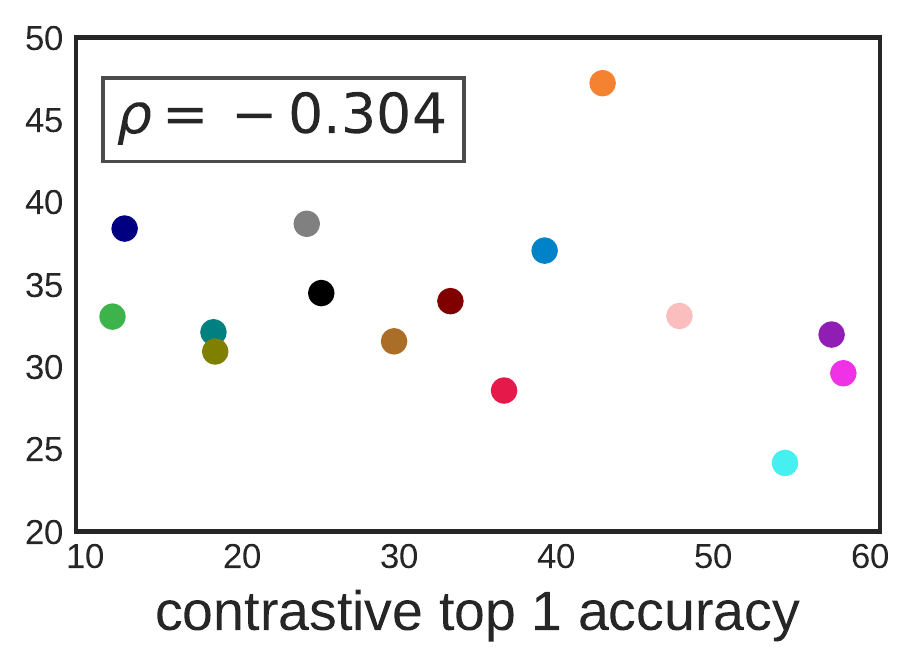}
\end{subfigure}
\begin{subfigure}{.28\textwidth}
    \centering
    \includegraphics[scale=0.48]{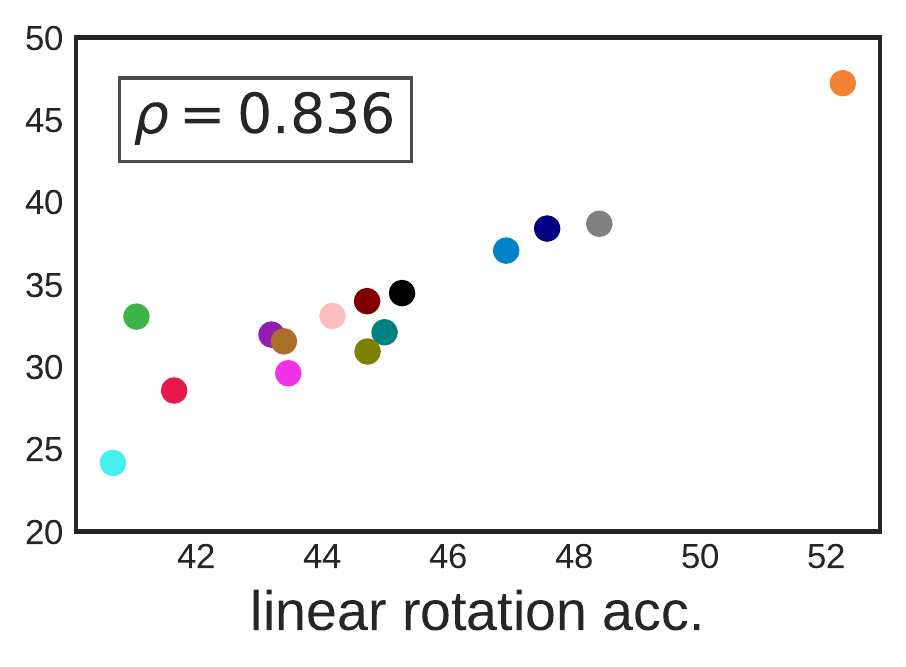}
\end{subfigure}
\begin{subfigure}{.1\textwidth}
    \centering
    \includegraphics[scale=0.4]{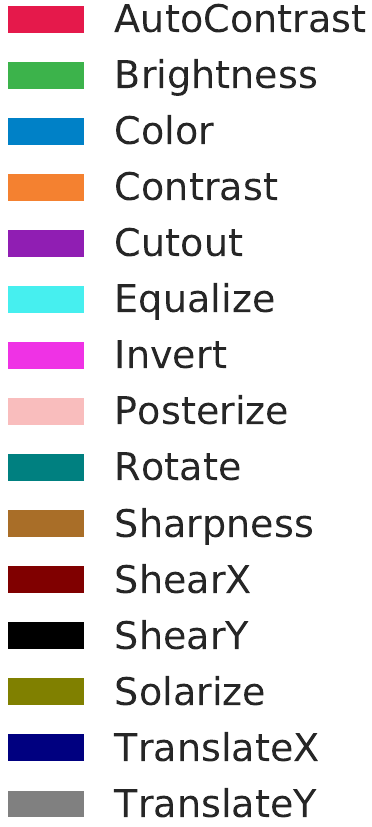}
\end{subfigure}
 
\caption[short]{For SVHN, we plot the supervised classification accuracy (y-axis) vs the InfoNCE loss function (left), contrastive top-1 accuracy (middle), and self-supervised linear rotation accuracy (right), for a self-supervised model trained using one of each transformation used by SelfAugment. Neither of the left two training metrics are a consistent measure of the quality of the representations, while the rotation prediction accuracy provides a strong linear relationship.}
\label{fig:single-aug}
\end{figure*}

\textbf{Linear Evaluation:} We first compare the rotation, jigsaw, and colorization evaluation tasks on the models obtained by MoCo pre-training with the above augmentation policies (full details in Appendix \ref{a:expdetails}). We compute the Spearman rank correlation~\cite{spearman1904proof} between the top-1 supervised linear classification accuracy and each evaluation task, where a higher correlation indicates a better evaluation task. As shown below, the rotation prediction task has a uniformly higher correlation with the supervised linear classification compared to the jigsaw and colorization tasks:
\begin{table}[h!]
\small
  \label{table:other-self-sup}
  \centering
  \begin{tabular}{rccc}
Evaluation & CIFAR-10 ($\rho$) & SVHN ($\rho$) & ImageNet ($\rho$)\\
\hline
Rotation & \textbf{0.966} & \textbf{0.948} & \textbf{0.986} \\
Jigsaw & 0.919 & 0.904  & 0.881 \\
Colorization & 0.612 & 0.806 & 0.627 \\
  \end{tabular}
\end{table}

The rotation evaluation correlations indicate a very strong relationship with the supervised evaluations, and based on its improvement over the jigsaw and colorization evaluations, we focus our main experiments, ablations, and SelfAugment/SelfRandAugment implementation of the automatic augmentation algorithms on this evaluation task. Appendix \ref{a:detailcorr} and \ref{a:rvsim} contain further details and analyses between the self-supervised evaluation tasks, where for instance, we also observe that the network activations from the rotation layer are more similar to the activations from the supervised classification layer compared to the other evaluations. We note that a rotation-based evaluation will not work for rotation invariant images (similarly, a color-based evaluation will not work for black-and-white images), and discuss this direction for future work in Appendix~\ref{a:detailcorr}.

The top row of Figure~\ref{fig:corrstudy} shows scatterplots of the correlation between the supervised and self-supervised evaluations for the rotation evaluation task, where the poorer performing models come from single transformation augmentation policies and early evaluation schedules, and the better performing models come from the RandAugment, SelfAugment, and MoCoV2 augmentation policies. We observe that the rank correlation is maintained for both the poor and strong performing models, indicating that the rotation evaluation can be used across a wide range of model performance.

\textbf{Transfer Learning:} A central goal of representation learning is to learn transferable features. We therefore study the correlation between rotation evaluation and the ImageNet transfer performance for the following datasets/tasks:
\begin{itemize}[leftmargin=*]
    \item \textbf{VOC07}~\cite{pascal-voc-2007} \textbf{object detection}: Following the specifics from \cite{he2019momentum}, we transfer the pre-trained models to a Faster R-CNN R50-C4 model and fine-tune all layers. Over three runs, we evaluate the mean results on VOC07 using the challenging COCO metric, AP$_{50:95}$, and report these results in Table~\ref{table:results}. Further, we report the AP$_{50}$/AP$_{75}$ in Appendix~\ref{a:detailcorr}. Fine-tuning is performed on the \texttt{train2007+2012} set and evaluation is on the \texttt{test2007} set.
    
    \item \textbf{COCO2014}~\cite{lin2014microsoft} \textbf{multi-class image classification} Following \cite{goyal2019scaling}, we train linear SVMs~\cite{boser1992training} on the frozen network and evaluate the accuracy over three end-to-end runs, denoted as COCO-C. \textbf{instance segmentation}, We use Mask-RCNN~\cite{he2017mask} with R50-FPN~\cite{lin2017feature} as our base model and add new Batch Normalization layers before the FPN parameters. Unlike the classification, training is performed on~\texttt{train2017} split with ${\sim}$118k images, and testing is performed on~\texttt{val2017} split. We report the Average Precision on masks (AP$^{\text{mk}}$) for the standard 1x schedule, denoted as COCO-mk.
    
    \item \textbf{Places205}~\cite{zhou2014learning} \textbf{low shot scene classification:} Following \cite{goyal2019scaling}, we train linear SVMs on the frozen network using $k=\{1, 4, 8, 16, 32, 64\}$ labeled examples and evaluate the accuracy over five runs, with the average across all $k$ used as the evaluation criteria, see Appendix~\ref{a:detailcorr} for a breakdown.
\end{itemize}

For VOC07, COCO2014, and Places205, the bottom row of Figure~\ref{fig:corrstudy} shows the ImageNet rotation performance vs the transfer task performance, yielding strong rank correlations of $\rho=\{0.968, 0.988, 0.982\}$, respectively. For comparison, the rank correlation of the supervised linear classification on ImageNet is $\rho_s=\{0.936, 0.952, 0.947\}$. For each transfer task, the rotation correlation is stronger than the supervised correlation. In \cite{chen_improved_2020}, the authors observe that ``linear classification accuracy is not monotonically related to transfer performance
in detection,'' an observation that we find further evidence for across more transfer tasks. Furthermore, we observe that rotation prediction has a stronger transfer correlation than linear classification.

\textbf{Finding the best individual image transformations} Similar to the exhaustive evaluation of single-transform augmentation policies performed in \cite{chen_simple_2020}, Figure~\ref{fig:single-aug} shows the performance of single-transform policies for SVHN (additional datasets in Appendix~\ref{a:detailcorr}). The left and middle plots show the supervised classification accuracy compared with the InfoNCE loss and top-1 contrastive accuracy (how well the instance contrastive model predicts the augmented image pairs), while the right plot shows the rotation prediction accuracy for the image transformations in $\mathbb{O}$ evaluated after $100$ training epochs. Using high or low values of InfoNCE or contrastive accuracy to select the best transformations would select a mixture of mediocre transformations,  missing the top performing transformation in the middle. 
By using the rotation prediction, each transformation has a clear linear relationship with the supervised performance, enabling the unsupervised selection of the best transformations.
 
\begin{figure}[hbt]
\centering
    \includegraphics[scale=0.6]{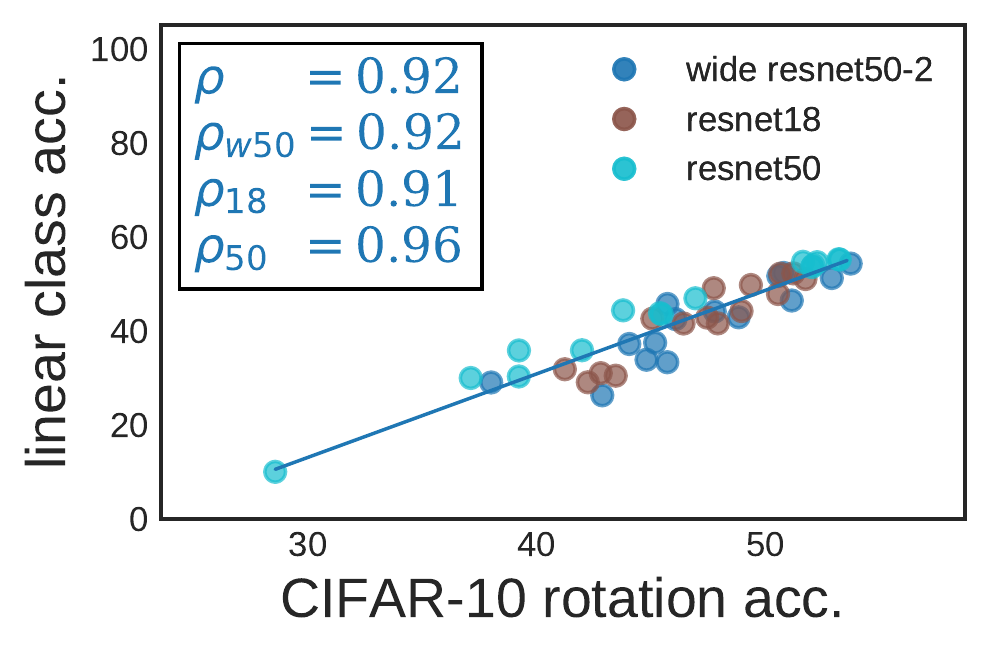}
\caption[short]{The Spearman rank correlation for CIFAR-10 RandAugment grid search for ResNet18, ResNet50, and Wide-ResNet-50-2, as well as the combined rank correlation.}
\label{fig:corr-arch}
\end{figure}

\textbf{Selecting augmentation policies across architectures} In \cite{kolesnikov2019revisiting}, the authors showed that when training an entire network to classify image rotations, rather than just a linear layer, the rotation prediction performance did not correlate across architectures. We study this same problem but using only a linear evaluation layer. Specifically, for CIFAR-10 we study the rotation and supervised evaluation correlation for ResNet18, ResNet50, Wide-ResNet-50-2, using RandAugment with a grid search over the number of transformations $N_\tau=\{1,2,3\}$ and magnitudes $\lambda=\{4,5,7,9,11\}$ evaluated after 100 epochs.
Figure~\ref{fig:corr-arch} shows the results for each architecture: the overall Spearman rank correlation across all architectures is $\rho=0.921$, between the Wide-ResNet-50-2 and ResNet18 Spearman correlations of $0.924$ and $0.914$ and less than the ResNet50 correlation of $0.957$.

\begin{table*} 
\small
  \caption{Top-1 accuracy of SelfAug on CIFAR-10, SVHN, ImageNet, as well as ImageNet transfer tasks and semi-supervised experiments. An ``R'' superscript denotes best rotation accuracy. COCO-c denotes multi-label image classification on COCO2014 and COCO0-mk denotes mask-RCCN instance segmentation on COCO2017. Bold results are greater than one standard deviation better across multiple runs, see Appendix~\ref{a:expdetails}. Without using labels or hand-tuning hyperparameters, SelfAugment results in better performance than MoCoV2 for 2 out of 3 benchmark datasets it was directly trained on and comparable transfer performance.}
  \label{table:results}
  \centering
  \begin{tabular}{llllllllll}
    & \multicolumn{3}{c}{Self-Supervised}     
    & \multicolumn{4}{c}{Transfer}  
    & \multicolumn{2}{c}{Semi-Supervised}  \\
    \cmidrule(lr){2-4}\cmidrule(lr){5-8} \cmidrule{9-10}
    \hline
      \textbf{unsup. feedback} & C10 & SVHN & IN & VOC & COCO-c & COCO-mk & Places & IN-1\% & IN-10\%\\
      
    \hline
    Base Aug & 89.1 & 89.2 & 46.7 & 47.1 & 33.6 & 29.9 & 5.4 & 16.0 & 33.6    \\
    SelfRandAug &90.3 & \textbf{96.8}$^R$  & 64.1 & 53.1 & 58.4 & 33.8 & 19.8 &36.1 &53.4   \\    
    SelfAug (min rot) & 91.0 & 94.9   & 57.4 & 50.2 & 50.9 & 31.9 &13.6 &26.4 & 45.1  \\
    SelfAug (min Info) & 87.5 & 86.0  & 51.7 & 49.2 &44.80 & 31.2 & 9.9 &20.5 & 38.6  \\
    SelfAug (max Info)  & 90.1 & 96.2 & 63.3 & 52.6 & 57.8 & 33.8 & 19.4 & 34.4 &52.7  \\
    SelfAug (minimax) & \textbf{92.6}$^R$ & 95.8 & \textbf{64.4} & 53.0 & 58.7 & 34.4 & \textbf{21.2} & 36.2 & 54.1 \\
    \hline
        \multicolumn{5}{l}{\textbf{supervised feedback}} \\
    \hline
    MoCoV2 \cite{chen_improved_2020} & 92.3 & 96.4 & 64.2$^R$ & \textbf{54.0} & \textbf{59.6} & 34.5 & 20.8 &\textbf{37.9}$^R$ &\textbf{54.9}$^R$  \\
    \hline
  \end{tabular}
\end{table*}

Overall, these strong correlations indicate that a linear rotation prediction evaluation is effective across architectures. In Appendix~\ref{a:detailcorrmlp}, we show that a drop in correlation occurs when using a two-layer MLP for rotation evaluation instead of a linear layer. Combined with the lack of correlation discovered when using a full network in \cite{kolesnikov2019revisiting}, we surmise that using a \emph{linear evaluation layer} to classify rotations is important as it disentangles the learned representations from the ability of the network to learn its own rotation features.

\subsection{Performance benchmarks} 

Here, we evaluate the SelfAugment and SelfRandAugment augmentation policies with the goal of establishing comparable results to the state-of-the art policies obtained through supervised feedback. 

\textbf{Setup:} We evaluate: \textbf{(i)} linear classification performance on top of the frozen network using CIFAR-10, SVHN, and ImageNet, \textbf{(ii)} transfer performance of ImageNet models on PASCAL VOC07 object detection, COCO2014 multi-class image classification, and Places205 few-show scene classification as described in the previous subsection, \textbf{(iii)} semi-supervised ImageNet top-1 accuracy with using only $1\%$ or $10\%$ of labels for linear training. All methods were evaluated with the same number of epochs and hyperparameters: 750 pre-train and 150 linear epochs for CIFAR-10/SVHN, 100 pre-train and 50 linear epochs for ImageNet, 24k fine-tuning iterations on VOC07, and the exact training parameters/schedules from \cite{goyal2019scaling} for COCO2014 and Places205. For SelfAugment, we used the settings from \cite{lim_fast_2019}: $P=10$ policies, $T=2$ transformations, and $K=5$ folds of training. For SelfRandAugment, we used the grid search described in the previous subsection. See Appendix~\ref{a:expdetails} for all details.  

\textbf{Linear classification:} Table~\ref{table:results} compares all versions of SelfAugment to the MoCoV2 augmentation policies~\cite{chen_improved_2020} that were based on the extensive study of supervised policy evaluations in \cite{chen_simple_2020}. 
For linear classification with CIFAR-10, SVHN, and ImageNet, SelfAugment  policies outperform  MoCoV2's policy. Where the largest gain occurs in the SVHN dataset. SVHN, consisting of images of house numbers, is the most distinct from ImageNet's diverse, object centric images. Since MoCoV2's policy is the result of extensive, supervised study on ImageNet, it does not transfer as well to a distributionally distinct dataset such as SVHN. These results indicate that SelfAugment is a stronger approach to obtaining quality representations for datasets that are distributionally distinct from ImageNet.

\textbf{Transfer and semi-supervised:} For the VOC07 object detection and COCO2014 image classification transfer tasks, the MoCoV2 policy performed best. For COCO2017 instance segmentation, SelfAugment and MoCoV2 had similar transfer performance (0.1 mask AP difference), while SelfAugment had a stronger transfer performance for few-shot scene classification on the Places205 dataset at each $k$ value (see Appendix~\ref{a:expdetails} for all details). Like ImageNet, VOC07 and COCO are natural images containing objects, while Places205 is a scene classification benchmark, consisting of complex scenes and diverse settings. SelfAugment's policies, as with the linear classification, perform better for the dataset and task that substantially differs from ImageNet. 

While the SelfAugment policy outperformed MoCoV2 for the linear classification with 100\% of the labels used for training, MoCoV2 performed better when using only $1\%$ and $10\%$ of the labeled data for training the linear classifier. These results indicate that using supervised evaluation to select a policy can lead to strong semi-supervised performance on the same dataset, but as indicated by the Places205 results, this policy may not transfer to other semi-supervised tasks.

\textbf{Rotation prediction:} The mean rank correlation for the supervised linear classification and rotation prediction for all results in this subsection is $\rho=0.978$, indicating that the strong correlation holds when removing the poorer performing models from the previous subsection. For every linear classification result except ImageNet, the top performing augmentation policy also corresponds to the top performing rotation prediction performance, as indicated with an ``R'' superscript. For ImageNet, MoCoV2's rotation prediction performance was significantly better than all other policies: $76.7\pm0.2\%$ compared to $73.6\pm0.2\%$ for SelfAugment's minimax, over three linear evaluations. While for a similar analysis, the linear classification performance for MoCoV2 performed worse: $64.2\pm0.1\%$ compared to $64.4\pm0.1\%$. 

As shown in the previous subsection, however, the rotation prediction has a stronger correlation to transfer performance than the supervised linear classification, and indeed, that is the case here: on the VOC07 and COCO2014 transfer tasks and the ImageNet 1\% and 10\% semi-supervised evaluations, the MoCoV2 policy significantly outperformed all other policies, while the SelfAugment minimax policy had the best transfer performance for the Places205 task. In other words, \emph{across the transfer and semi-supervised evaluations, rotation prediction was more indicative of performance than supervised linear classification.}

\section{Discussion}
\label{discussion}

We have shown that a self-supervised rotation evaluation has a strong correlation with supervised evaluation (outperforming jigsaw/colorization tasks) and that this evaluation can be incorporated into an effective loss function for efficient augmentation selection ($\S\ref{experiments}$). Here, we further reflect on the utility of the self-supervised rotation evaluation task.

\textbf{If rotation prediction is highly correlated with supervised evaluations, why not directly train on it?}
As discussed in Appendix \ref{a:detailcorrmlp}, the authors of \cite{kolesnikov2019revisiting} found that rotation accuracy from training a full network on rotation prediction was only weakly correlated with supervised performance. Furthermore, as discussed in Appendix \ref{a:detailcorrmlp}, using a 2-layer MLP for rotation prediction drops the correlation from $\rho = $.966 to .904 even though the prediction accuracy improves. This indicates that while rotation prediction using a linear combination of the learned representations is an effective \emph{evaluation}, actually learning the representations via rotation prediction is not as strongly correlated.

\textbf{Shouldn't minimizing rotation error yield the best policies?}
We evaluate policies only \emph{after} contrastive training, so minimizing rotation error yields policies that improve rotation prediction if we were to retrain the linear classifier. However, as indicated by the poor performance of using the rotation-error as the loss function, it is important to also take the contrastive task into consideration when retraining the entire network. To explore this idea further, we minimize a supervised classification loss function for SelfAugment. 
Due to the strong correlation between the rotation task and supervised evaluation, this results in similar performance to minimizing rotation loss with SelfAugment: we observe an accuracy of 90.7 on CIFAR-10 and 94.9 on SVHN using supervised feedback (rotation evaluation yields an accuracy of 91.0 on CIFAR-10 and 93.7 on SVHN). This performance is worse than the loss functions that simultaneously maximize the InfoNCE loss, which encourage difficult augmentation policies (see Appendix~\ref{a:lossfun}).

\textbf{Which SelfAugment loss function should be used on a new, unlabeled dataset?} 
When training with a new unlabeled dataset, we recommend starting with the minimax SelfAugment loss function for augmentation policy selection.
This recommendation stems from its superior performance across all datasets and tasks. For comparison, the SelfAugment loss function minimizing the InfoNCE produced results that were often worse than the baseline policy, while both maximizing the InfoNCE or minimizing rotation prediction exceeded the baseline, but generally did not surpass MoCoV2's policy. In Appendix~\ref{a:lossfun}, we show the effective magnitude of each individual transformation for each loss functions. We see that, as expected, the minimax loss function produces policies with intermediate  magnitudes across all datasets. Its strong performance indicate that these intermediate magnitudes have found a \emph{sweet spot} where the augmentations strike a balance between creating difficult instance contrastive tasks and retaining salient image features.

It is worth noting that across both SVHN and CIFAR-10, SelfAugment with min InfoNCE loss function results in \emph{worse} results than simply using the baseline augmentation. We believe this is because the selected augmentations make it easier to tell samples apart using weak augmentations that do not actually add meaningful contrastive tasks (see Appendix~\ref{a:lossfun}). These augmentations minimize InfoNCE but do not appear to be helpful in improving the model's ability to generalize to new data. Accordingly, the model performs poorly when retrained with these augmentations. 

\textbf{Computational efficiency:} 
Importantly, SelfAugment and RandAugment provide an efficient framework for finding augmentations for instance contrastive learning without using labels. Prior work~\cite{chen_simple_2020} performed grid searches over various augmentation policies, requires training many models to completion. Training a single instance of MoCoV2 on ImageNet using 4 V100 GPUs for 100 epochs takes 52 hours (208 GPU hours). SelfAugment finds useful, and typically improved, augmentation policies in 730 GPU hours, and can learn additional augmentation policies in an additional 98 GPU Hours (see Appendix \ref{a:eff} for details). This is less than the computational cost of training just four total instances. For comparison, \cite{chen_simple_2020} reported at least 49 full training runs. 

\section{Conclusion}
In this paper, we identified the problem that self-supervised representations are evaluated using labeled data, oftentimes using the labels from the ``unlabeled'' training dataset itself. We established that a self-supervised image rotation task is strongly correlated with the supervised performance for standard computer vision recognition tasks, and as a result, can be used to evaluate learned representations. Using this evaluation, we establish two unsupervised data augmentation policy selection algorithms and show that they can outperform or perform comparably to policies obtained using supervised feedback.

\textbf{Acknowledgements}
Prof. Darrell’s group was supported in part by DoD including DARPA's XAI, LwLL, and/or SemaFor programs, as well as Berkeley's BAIR industrial alliance programs. Prof. Keutzer's group was supported in part by Alibaba, Amazon, Google, Facebook, Intel, and Samsung as well as BAIR and BDD at Berkeley. We would additionally like to thank Pieter Abbeel, Tete Xiao, Roi Herzig, and Amir Bar for helpful feedback and discussions. Wandb \cite{wandb} graciously provided academic research accounts to track our experimental analyses.
{\small
\bibliographystyle{unsrt}
\bibliography{references}

\begin{thebibliography}{10}

\bibitem{chen_simple_2020}
Ting Chen, Simon Kornblith, Mohammad Norouzi, and Geoffrey Hinton.
\newblock A simple framework for contrastive learning of visual
  representations.
\newblock {\em arXiv preprint arXiv:2002.05709}, 2020.

\bibitem{henaff2019data}
Olivier~J H{\'e}naff, Aravind Srinivas, Jeffrey De~Fauw, Ali Razavi, Carl
  Doersch, SM~Eslami, and Aaron van~den Oord.
\newblock Data-efficient image recognition with contrastive predictive coding.
\newblock {\em arXiv preprint arXiv:1905.09272}, 2019.

\bibitem{he2019momentum}
Kaiming He, Haoqi Fan, Yuxin Wu, Saining Xie, and Ross Girshick.
\newblock Momentum contrast for unsupervised visual representation learning.
\newblock In {\em Proceedings of the IEEE Conference on Computer Vision and
  Pattern Recognition}, 2020.

\bibitem{chen_improved_2020}
Xinlei Chen, Haoqi Fan, Ross Girshick, and Kaiming He.
\newblock Improved baselines with momentum contrastive learning.
\newblock {\em arXiv preprint arXiv:2003.04297}, 2020.

\bibitem{wu2018unsupervised}
Zhirong Wu, Yuanjun Xiong, Stella~X Yu, and Dahua Lin.
\newblock Unsupervised feature learning via non-parametric instance
  discrimination.
\newblock In {\em Proceedings of the IEEE Conference on Computer Vision and
  Pattern Recognition}, pages 3733--3742, 2018.

\bibitem{tian2020makes}
Yonglong Tian, Chen Sun, Ben Poole, Dilip Krishnan, Cordelia Schmid, and
  Phillip Isola.
\newblock What makes for good views for contrastive learning?
\newblock {\em arXiv preprint arXiv:2005.10243}, 2020.

\bibitem{medical}
H.~{Shin}, M.~{Orton}, D.~J. {Collins}, S.~{Doran}, and M.~O. {Leach}.
\newblock Autoencoder in time-series analysis for unsupervised tissues
  characterisation in a large unlabelled medical image dataset.
\newblock In {\em 2011 10th International Conference on Machine Learning and
  Applications and Workshops}, volume~1, pages 259--264, 2011.

\bibitem{zhang2017imaterialist}
X~Zhang, Y~Cui, Y~Song, H~Adam, and S~Belongie.
\newblock The imaterialist challenge 2017 dataset.
\newblock In {\em FGVC Workshop at CVPR}, volume~2, page~3, 2017.

\bibitem{yang2018end}
Zhengyuan Yang, Yixuan Zhang, Jerry Yu, Junjie Cai, and Jiebo Luo.
\newblock End-to-end multi-modal multi-task vehicle control for self-driving
  cars with visual perceptions.
\newblock In {\em 2018 24th International Conference on Pattern Recognition
  (ICPR)}, pages 2289--2294. IEEE, 2018.

\bibitem{c10}
Alex Krizhevsky, Vinod Nair, and Geoffrey Hinton.
\newblock Cifar-10 (canadian institute for advanced research).

\bibitem{goodfellow2013multi}
Ian~J Goodfellow, Yaroslav Bulatov, Julian Ibarz, Sacha Arnoud, and Vinay Shet.
\newblock Multi-digit number recognition from street view imagery using deep
  convolutional neural networks.
\newblock {\em arXiv preprint arXiv:1312.6082}, 2013.

\bibitem{ILSVRC15}
Olga Russakovsky, Jia Deng, Hao Su, Jonathan Krause, Sanjeev Satheesh, Sean Ma,
  Zhiheng Huang, Andrej Karpathy, Aditya Khosla, Michael Bernstein,
  Alexander~C. Berg, and Li~Fei-Fei.
\newblock {ImageNet Large Scale Visual Recognition Challenge}.
\newblock {\em International Journal of Computer Vision (IJCV)},
  115(3):211--252, 2015.

\bibitem{pascal-voc-2007}
M.~Everingham, L.~Van~Gool, C.~K.~I. Williams, J.~Winn, and A.~Zisserman.
\newblock The {PASCAL} {V}isual {O}bject {C}lasses {C}hallenge 2007 {(VOC2007)}
  {R}esults.
\newblock
  http://www.pascal-network.org/challenges/VOC/voc2007/workshop/index.html.

\bibitem{lin2014microsoft}
Tsung-Yi Lin, Michael Maire, Serge Belongie, James Hays, Pietro Perona, Deva
  Ramanan, Piotr Doll{\'a}r, and C~Lawrence Zitnick.
\newblock Microsoft coco: Common objects in context.
\newblock In {\em European conference on computer vision}, pages 740--755.
  Springer, 2014.

\bibitem{zhou2014learning}
Bolei Zhou, Agata Lapedriza, Jianxiong Xiao, Antonio Torralba, and Aude Oliva.
\newblock Learning deep features for scene recognition using places database.
\newblock In {\em Advances in neural information processing systems}, pages
  487--495, 2014.

\bibitem{noroozi2016unsupervised}
Mehdi Noroozi and Paolo Favaro.
\newblock Unsupervised learning of visual representations by solving jigsaw
  puzzles.
\newblock In {\em European Conference on Computer Vision}, pages 69--84.
  Springer, 2016.

\bibitem{zhang2016colorful}
Richard Zhang, Phillip Isola, and Alexei~A Efros.
\newblock Colorful image colorization.
\newblock In {\em European conference on computer vision}, pages 649--666.
  Springer, 2016.

\bibitem{reed2021self}
Colorado~J Reed, Xiangyu Yue, Ani Nrusimha, Sayna Ebrahimi, Vivek Vijaykumar,
  Richard Mao, Bo~Li, Shanghang Zhang, Devin Guillory, Sean Metzger, et~al.
\newblock Self-supervised pretraining improves self-supervised pretraining.
\newblock {\em arXiv preprint arXiv:2103.12718}, 2021.

\bibitem{donahue2014decaf}
Jeff Donahue, Yangqing Jia, Oriol Vinyals, Judy Hoffman, Ning Zhang, Eric
  Tzeng, and Trevor Darrell.
\newblock Decaf: A deep convolutional activation feature for generic visual
  recognition.
\newblock In {\em International conference on machine learning}, pages
  647--655, 2014.

\bibitem{erhan2010does}
Dumitru Erhan, Yoshua Bengio, Aaron Courville, Pierre-Antoine Manzagol, Pascal
  Vincent, and Samy Bengio.
\newblock Why does unsupervised pre-training help deep learning?
\newblock {\em Journal of Machine Learning Research}, 11(Feb):625--660, 2010.

\bibitem{zeiler2014visualizing}
Matthew~D Zeiler and Rob Fergus.
\newblock Visualizing and understanding convolutional networks.
\newblock In {\em European conference on computer vision}, pages 818--833.
  Springer, 2014.

\bibitem{goodfellow2016deep}
Ian Goodfellow, Yoshua Bengio, and Aaron Courville.
\newblock {\em Deep learning}.
\newblock MIT press, 2016.

\bibitem{lecun2015deep}
Yann LeCun, Yoshua Bengio, and Geoffrey Hinton.
\newblock Deep learning.
\newblock {\em nature}, 521(7553):436--444, 2015.

\bibitem{radford2018improving}
Alec Radford, Karthik Narasimhan, Tim Salimans, and Ilya Sutskever.
\newblock Improving language understanding by generative pre-training.

\bibitem{devlin2018bert}
Jacob Devlin, Ming-Wei Chang, Kenton Lee, and Kristina Toutanova.
\newblock Bert: Pre-training of deep bidirectional transformers for language
  understanding.
\newblock {\em arXiv preprint arXiv:1810.04805}, 2018.

\bibitem{oord2018representation}
Aaron van~den Oord, Yazhe Li, and Oriol Vinyals.
\newblock Representation learning with contrastive predictive coding.
\newblock {\em arXiv preprint arXiv:1807.03748}, 2018.

\bibitem{coates2012learning}
Adam Coates and Andrew~Y Ng.
\newblock Learning feature representations with k-means.
\newblock In {\em Neural networks: Tricks of the trade}, pages 561--580.
  Springer, 2012.

\bibitem{rotation}
Spyros Gidaris, Praveer Singh, and Nikos Komodakis.
\newblock Unsupervised representation learning by predicting image rotations.
\newblock {\em CoRR}, abs/1803.07728, 2018.

\bibitem{doersch2015unsupervised}
Carl Doersch, Abhinav Gupta, and Alexei~A Efros.
\newblock Unsupervised visual representation learning by context prediction.
\newblock In {\em Proceedings of the IEEE International Conference on Computer
  Vision}, pages 1422--1430, 2015.

\bibitem{gidaris2018unsupervised}
Spyros Gidaris, Praveer Singh, and Nikos Komodakis.
\newblock Unsupervised representation learning by predicting image rotations,
  2018.

\bibitem{kolesnikov2019revisiting}
Alexander Kolesnikov, Xiaohua Zhai, and Lucas Beyer.
\newblock Revisiting self-supervised visual representation learning.
\newblock In {\em Proceedings of the IEEE conference on Computer Vision and
  Pattern Recognition}, pages 1920--1929, 2019.

\bibitem{puigcerver2020scalable}
Joan Puigcerver, Carlos Riquelme, Basil Mustafa, Cedric Renggli,
  Andr{\'e}~Susano Pinto, Sylvain Gelly, Daniel Keysers, and Neil Houlsby.
\newblock Scalable transfer learning with expert models.
\newblock {\em arXiv preprint arXiv:2009.13239}, 2020.

\bibitem{renggli2020model}
Cedric Renggli, Andr{\'e}~Susano Pinto, Luka Rimanic, Joan Puigcerver, Carlos
  Riquelme, Ce~Zhang, and Mario Lucic.
\newblock Which model to transfer? finding the needle in the growing haystack.
\newblock {\em arXiv preprint arXiv:2010.06402}, 2020.

\bibitem{shorten_survey_2019}
Connor Shorten and Taghi~M. Khoshgoftaar.
\newblock A survey on {Image} {Data} {Augmentation} for {Deep} {Learning}.
\newblock {\em Journal of Big Data}, 6(1):60, July 2019.

\bibitem{cubuk_autoaugment_2019}
Ekin~D Cubuk, Barret Zoph, Dandelion Mane, Vijay Vasudevan, and Quoc~V Le.
\newblock Autoaugment: Learning augmentation strategies from data.
\newblock In {\em Proceedings of the IEEE conference on computer vision and
  pattern recognition}, pages 113--123, 2019.

\bibitem{ho_population_2019}
Daniel Ho, Eric Liang, Ion Stoica, Pieter Abbeel, and Xi~Chen.
\newblock Population based augmentation: Efficient learning of augmentation
  policy schedules.
\newblock In {\em ICML}, 2019.

\bibitem{lim_fast_2019}
Sungbin Lim, Ildoo Kim, Taesup Kim, Chiheon Kim, and Sungwoong Kim.
\newblock Fast autoaugment.
\newblock In {\em Advances in Neural Information Processing Systems}, pages
  6662--6672, 2019.

\bibitem{cubuk_randaugment_2019}
Ekin~D Cubuk, Barret Zoph, Jonathon Shlens, and Quoc~V Le.
\newblock Randaugment: Practical data augmentation with no separate search.
\newblock {\em arXiv preprint arXiv:1909.13719}, 2019.

\bibitem{liu2020labels}
Chenxi Liu, Piotr Doll{\'a}r, Kaiming He, Ross Girshick, Alan Yuille, and
  Saining Xie.
\newblock Are labels necessary for neural architecture search?
\newblock {\em arXiv preprint arXiv:2003.12056}, 2020.

\bibitem{szegedy2017inception}
Christian Szegedy, Sergey Ioffe, Vincent Vanhoucke, and Alexander~A Alemi.
\newblock Inception-v4, inception-resnet and the impact of residual connections
  on learning.
\newblock In {\em Thirty-first AAAI conference on artificial intelligence},
  2017.

\bibitem{spearman1904proof}
C~Spearman.
\newblock The proof and measurement of association between two things.
\newblock {\em The American Journal of Psychology}, page~72, 1904.

\bibitem{goyal2019scaling}
Priya Goyal, Dhruv Mahajan, Abhinav Gupta, and Ishan Misra.
\newblock Scaling and benchmarking self-supervised visual representation
  learning.
\newblock In {\em Proceedings of the IEEE International Conference on Computer
  Vision}, pages 6391--6400, 2019.

\bibitem{boser1992training}
Bernhard~E Boser, Isabelle~M Guyon, and Vladimir~N Vapnik.
\newblock A training algorithm for optimal margin classifiers.
\newblock In {\em Proceedings of the fifth annual workshop on Computational
  learning theory}, pages 144--152, 1992.

\bibitem{he2017mask}
Kaiming He, Georgia Gkioxari, Piotr Doll{\'a}r, and Ross Girshick.
\newblock Mask r-cnn.
\newblock In {\em Proceedings of the IEEE international conference on computer
  vision}, pages 2961--2969, 2017.

\bibitem{lin2017feature}
Tsung-Yi Lin, Piotr Doll{\'a}r, Ross Girshick, Kaiming He, Bharath Hariharan,
  and Serge Belongie.
\newblock Feature pyramid networks for object detection.
\newblock In {\em Proceedings of the IEEE conference on computer vision and
  pattern recognition}, pages 2117--2125, 2017.

\bibitem{wandb}
Lukas Biewald.
\newblock Experiment tracking with weights and biases, 2020.
\newblock Software available from wandb.com.

\bibitem{similarity}
Jessica A.~F. Thompson, Yoshua Bengio, and Marc Sch{\"{o}}nwiesner.
\newblock The effect of task and training on intermediate representations in
  convolutional neural networks revealed with modified {RV} similarity
  analysis.
\newblock {\em CoRR}, abs/1912.02260, 2019.

\end{thebibliography}
}

\clearpage

\appendix

\clearpage

\section{Notation and definitions}
\label{a:notation}
\begin{table}[h!]
    \centering
    \begin{tabular}{lp{5cm}c}
\toprule
\textbf{Notation} & \textbf{Definition} \\
\hline
$\rho$ & Spearman rank correlation \\
\hline
$\theta_{moco}$ & MoCo encoder \\
\hline
$\phi_{ss}$ & Linear self-supervised evaluation head \\
\hline
$D$ & Dataset \\
\hline
$K$ & Number of folds used for training SelfAugment \\
\hline
$T$ & Number of augmentations to apply to each image at each iteration \\
\hline
$B$ &  Number of policies to consider during Bayesian optimization search\\
\hline
$P$ &  Top number of policies to select from each fold in SelfAugment\\ 
\hline
$\mathbb{O}$ & The set of candidate image transformations for an augmentation policy \\
\hline
$\mathcal{L}$ & A loss function \\
\hline
$\mathcal{O}$ & An image transformations in $\mathbb{O}$ \\
\hline
$\mathcal{S}$ & Set of sub-policies \\
\hline
$\mathcal{M}$ & A model (e.g. a neural network) \\
\hline
$\tau$ & $\tau\in\mathcal{S}$ is the sequential application of $N_{\tau}$ consecutive transformations \\
\hline
$N_\tau$ & The number of consecutive transformations to apply in a sub-policy \\
\hline
$\lambda$ & The magnitude of an image transformation \\
\hline
FAA & Fast AutoAugment, from \cite{lim_fast_2019} \\
\hline
InfoNCE & See Eq~\ref{eq:infonce_loss} \\
\hline
MoCo & Momentum Contrast, from \cite{he2019momentum} \\
\bottomrule
\end{tabular}
\end{table}

\section{Augmentation transformation details}
\begin{table*}[ht]
  \caption{PIL image transformations used for SelfAugment and RandAugment. The min and max magnitude values are taken and from \cite{cubuk_autoaugment_2019}}
  \label{table:aug}
  \centering
  \begin{tabular}{lp{6cm}cc}
  \hline
  \textbf{Name} &Description &Min ($\lambda = 0.0$) &Max ($\lambda = 1.0$) \\
\hline
ShearX &shear the image along the horizontal axis with magnitude rate &-0.3 & 0.3\\
\\[-0.5em]
ShearY &shear the image along the vertical axis with magnitude rate &-0.3 & 0.3\\
\\[-0.5em]
TranslateX &translate the image in the horizontal direction by magnitude percentage &-0.45 & 0.45\\
\\[-0.5em]
TranslateY &translate the image in the vertical direction by magnitude percentage  &-0.45 & 0.45\\
\\[-0.5em]
Rotate&rotate the image by magnitude degrees & -30 & 30\\
\\[-0.5em]
AutoContrast&adjust contrast so darkest pixel is black and lightest is white  & 0 & 1\\
\\[-0.5em]
Invert &invert the pixels of the image &0 & 1\\
\\[-0.5em] 
Solarize &invert the pixels above a magnitude threshold& 0 & 256\\
\\[-0.5em]
Posterize&reduce the number of bits for each color to magnitude & 4 & 8\\
\\[-0.5em]
Contrast &adjust image contrast, where magnitude 0 is grey and magnitude 1 is original image & 0.1 & 1.9\\
\\[-0.5em]
Color&adjust color of image such that magnitude 0 is black and white and magnitude 1 is original image &0.1&1.9\\
\\[-0.5em]
Brightness&brightness adjustment such that magnitude 0 is black image and 1 is original image&0.1&1.9\\
\\[-0.5em]
Sharpness&magnitude 0 is a blurred image and 1 is original image&0.1&1.9\\
\\[-0.5em]
Cutout&cutout a random square from the image with side length equal to the magnitude percentage of pixels&0&0.2\\
\\[-0.5em]
Equalize&equalize the image histogram&0&1\\
    \bottomrule
  \end{tabular}
\end{table*}

\label{a:transforms}

Following \cite{lim_fast_2019, cubuk_randaugment_2019}, we define the set of transformations used for SelfAugment and RandAugment, $\mathbb{O}$, as the PIL-based image transformations in Table~\ref{table:aug}. Each transformation has a minimum and maximum magnitude, $\lambda$, where for RandAugment, the entire range is discretized over $30$ integers. See \cite{cubuk_randaugment_2019, cubuk_autoaugment_2019} for further details and descriptions of each transformation.

\section{Additional training and experiment details}
\label{a:expdetails}

\textbf{CIFAR-10, SVHN, ImageNet}: Table \ref{tab:expdetails-params} lists the training and experimental parameters for CIFAR-10, SVHN, and ImageNet. The training parameters were taken from \cite{he2019momentum} and adjusted for 4 GPUs, i.e.~the learning rate and batch size were scaled by $0.5$ since \cite{he2019momentum} experiments were conducted on 8 GPUs. Consult \cite{he2019momentum} and \cite{chen_improved_2020} for MoCo parameter information. For the linear classifier, we used 50 training iterations where the learning rate was $10\times$ reduced at $30$ and $40$ epochs for ImageNet and $20$ and $30$ epochs for CIFAR-10 and SVHN. In \cite{he2019momentum}, the linear layer was trained over $150$ iterations with a reduction at $80$ and $100$ iterations. In early experiments, we found the performance converged much earlier, and to reduce computational resources, we reduced all linear training iterations. 

\textbf{VOC07} Following \cite{he2019momentum}, we transfer the ImageNet ResNet-50 weights to perform object detection using a Faster R-CNN R50-C4, with BN tuned. We fine-tuned all layers end-to-end. The image scale is [480, 800] pixels during training and 800 at inference. Training was on the VOC \texttt{trainval07+12} set and evaluation was on the \texttt{test2007} set. The R50-C4 backbones is similar to those available in Detectron2\footnote{\url{https://github.com/facebookresearch/detectron2}}, where the backbone stops at the conv4 stage, and the box prediction head consists of the conv5 stage followed by a BN layer. Table \ref{table:allap} displays the AP/AP$_{50}$/AP$_{75}$ breakdown for three fine-tunings.  

\textbf{COCO2014/Places205}: Following \cite{goyal2019scaling}, we train Linear SVMs on frozen feature representations. We train a linear SVM per class for (80 for COCO2014, 205 for Places205) for the cost values $C \in 2^{[-19, -4]} \cup  10^{[-7, 2]}$ We used 3-fold cross-validation to select the cost parameter per class and then further calculate the mean average precision. The features are first normalized in a (N, 9k) matrix, where N is number of samples in data and 9k is the resized feature dimension, to have norm=1 along each feature dimension. This normalization step is applied on evaluation data too. We use the following hyperparameter setting for training using \texttt{LinearSVC} sklearn class: \texttt{class\_weight} ratio of 2:1 for positive:negative classes, \texttt{penalty}=l2, \texttt{loss}=squared\_hinge, \texttt{tol}=1e-4, \texttt{dual}=True and \texttt{max\_iter}=2000. Table \ref{table:placesallk} displays the Places205 results across five linears SVM trainings for all $k$ values.

\begin{table*}
\centering
\caption{Transfer results: Places205 top-1 accuracy for frozen ResNet50 encoder after pre-training, with a linear SVM trained for scene classification, with $k=\{1,4,8,16,32,64\}$ labeled training images for each class, averaged over five SVM trainings, where the errors indicate the standard deviation (see text for details).}
\label{table:placesallk}
\begin{tabular}{lllllll}
\hline
\textbf{Labeled samples} & 1    & 4     & 8     & 16    & 32    & 64    \\
\hline
Base Aug             & $1.35\pm.03$ & $2.8\pm0.05$   & $4.18\pm.11$  & $5.88\pm.12$  & $8.04\pm.16$  & $10.13\pm.15$       \\
SelfRandAug          & $6.17\pm.05$ & $13.01\pm.15$ & $18.36\pm.12$ & $23.13\pm.11$ & $27.18\pm.11$ & $30.89\pm.09$ \\
SelfAug (min rot)    & $3.58\pm.06$ & $7.85\pm0.09$  & $11.77\pm.10$ & $15.60\pm.17$ & $19.66\pm.10$ & $23.26\pm.05$ \\
SelfAug (min Info)   & $2.46\pm.04$ & $5.36\pm0.07$  & $8.11\pm.11$  & $11.04\pm.07$ & $14.52\pm.09$ & $17.75\pm.29$ \\
SelfAug (max Info)   & $5.87\pm.06$ & $12.73\pm.14$ & $17.88\pm.16$ & $22.64\pm.21$ & $26.95\pm.12$ & $30.56\pm.17$ \\
\textbf{SelfAug (minimax)}    & $6.73\pm.08$ & $14.51\pm.20$ & $19.89\pm.18$ & $24.72\pm.10$ & $28.77\pm.13$ & $32.35\pm.09$ \\
MoCoV2               & $6.65\pm.11$ & $14.06\pm.13$ & $19.59\pm.14$ & $24.57\pm.13$ & $28.56\pm.08$ & $32.24\pm.10$ \\
\hline
\end{tabular}
\end{table*}

\begin{table*}
\centering
\caption{Transfer Result: For \texttt{VOC07 test}, this table reports the average AP50 and COCO-style AP/$\text{AP}_{75}$ over three runs of fine-tuning the ResNet50 encoder from ImageNet pre-training, where the errors indicate the standard deviation (see text for details).}
\label{table:allap}
\begin{tabular}{llll}
\hline
 \textbf{Evaluation}                  & AP    & AP$_{50}$ & AP$_{75}$ \\
\hline
Base Aug           & $47.08\pm 0.17$ & $74.80\pm 0.17$     & $50.26\pm 0.24$     \\
SelfRandAug        & $53.06\pm 0.21$ & $80.09\pm 0.14$     & $57.58\pm 0.29$     \\
SelfAug (min rot)  & $50.13\pm 0.20$ & $77.66\pm 0.18$     & $53.94\pm 0.21$     \\
SelfAug (min Info) & $49.18\pm 0.21$ & $76.31\pm 0.17$     & $53.17\pm 0.18$     \\
SelfAug (max Info) & $52.66\pm 0.18$ & $79.69\pm 0.15$     & $57.33\pm 0.25$     \\
SelfAug (minimax)  & $52.72\pm 0.22$ & $79.79\pm 0.21$     & $57.62\pm 0.27$     \\
\textbf{MoCoV2}    & $53.94\pm 0.23$ & $80.64\pm 0.18$     & $59.34\pm 0.31$ \\
\hline
\end{tabular} 
\end{table*}

\begin{table*}[t]

\centering
  \caption{Detailed training parameters for CIFAR-10, SVHN, and ImageNet experiments carried out in this paper.}
\begin{tabular}{lll}
\hline
-\textbf{MoCo Params} & \textbf{CIFAR 10/SVHN}  & \textbf{ImageNet}\\
\hline
 Batch Size & 512 & 128 \\
 moco-dim & 128 & 128 \\
 moco-k & 65536 & 65536 \\
 moco-m & 0.999 & 0.99 \\
 moco-t & 0.2 & 0.2 \\
 num-gpus & 4 & 4 \\
 lr & 0.4 & 0.015 \\
 schedule & 120, 160 & 60, 80 \\
 momentum & 0.9 & 0.9 \\
 weight decay & 1e-4 & 1e-4 \\
\hline
\textbf{Classifier Params} &  & \\
lr & 15 & 30 \\
batch size & 256 & 256 \\ 
momentum & 0.9  & 0.9 \\
weight decay & 0.0 & 0.0 \\
schedule & 20, 30 & 30, 40 \\
epochs & 50 & 50\\
\hline
\label{tab:expdetails-params}
\end{tabular}
\end{table*}

\clearpage
\clearpage

\section{Correlation study}
\label{a:detailcorr}
In this section, we detail the full experimental setup, investigation, and results from our study of the correlation between rotation prediction performance with a linear network and supervised downstream task performance. We also include additional preliminary and ablation studies.
\subsection{Correlation study details}
We study RandAugment, SelfAugment, and MoCoV2 augmentation policies, and then evaluate the performance using self-supervised rotation prediction with a linear layer. For CIFAR-10 and SVHN, we evaluate:
\begin{itemize}
    \item A \emph{base augmentation} of random left-right horizontal flips with $p=0.5$ of being applied and random resize and crop transformation with magnitude range $(0.2,1.0)$, trained for 750 epochs.
    \item On top of the base augmentation\footnote{Using this base augmentation systematically improved all RandAugment results over not using a base augmentation, both in terms of its rotation prediction performance and supervised linear classification performance.}, we performed a RandAugment grid search with magnitude $\lambda=\{4, 5, 7, 9, 11\}$ and number of transformations applied $N_\tau=\{1, 2, 3\}$, evaluated at $\{100, 500\}$ epochs. We initially included $N_\tau=4$ in the grid search, but this always led to a degenerate solution, whereby the training loss would not minimize the objective function and the evaluation would yield a chance result ($25\%$ self-supervised rotation prediction and $10\%$ supervised classification for CIFAR-10).
    \item Following \cite{cubuk_randaugment_2019}, we also experimented with scaling the magnitude parameter $\lambda$ from $[4,11]$ linearly throughout the training. We did this for each of the three $N_\tau$ values and evaluated the results at $500$ epochs.
    \item As part of the SelfAugment algorithm, we trained an augmentation policy consisting of each of the fifteen individual transformations in RandAugment (transformations listed in \S\ref{ss:self-supervised-data-aug}). In addition, we include the random resize crop transformation, as it was shown to be the best performing individual transformation in \cite{chen_simple_2020}. The magnitude for each transformation was stochastically selected from its magnitude range defined in \S\ref{a:transforms} at each iteration. Each of these 16 transformations were evaluated after a short training cycle of $100$ epochs. Each of these transformations were applied on top of a random left-right horizontal flip applied with $p=0.5$.
    \item Using the rotation prediction results from each of the individual transformation policies, we selected the top $K_T=\{3,6,9\}$ augmentations, and trained RandAugment using only these transformations. We performed this training for $\lambda=\{4, 7\}$ with $N_\tau=2$, evaluated at $500$ epochs.
    \item We further include the four SelfAugment policies from each of its loss functions, evaluated at $750$ epochs.
\end{itemize}
In total, this yields $61$ different models for each of CIFAR-10/SVHN. For ImageNet we evaluate:

\begin{itemize}
    \item A \emph{base augmentation} of random left-right horizontal flips with $p=0.5$ of being applied and random resize and crop transformation with magnitude range $(0.2,1.0)$, trained for 100 epochs.
    \item On top of the base augmentation\footnote{Using this base augmentation systematically improved all RandAugment results over not using a base augmentation, both in terms of its rotation prediction performance and supervised linear classification performance.}, we performed a RandAugment grid search with magnitude $\lambda=\{5, 7, 9, 11, 13\}$ and number of transformations applied $N_\tau=2$, evaluated at $\{20, 60, 100\}$ epochs.
    \item Following \cite{cubuk_randaugment_2019}, we also experimented with scaling the magnitude parameter $\lambda$ from $[5,13]$ linearly throughout the training. We did this for $N_\tau=2$ and evaluated the results at 100 epochs.
    \item As part of the SelfAugment algorithm, we trained an augmentation policy consisting of each of the fifteen individual transformations in RandAugment (transformations listed in \S\label{TODO}). In addition, we include the random resize crop transformation, as it was shown to be the best performing individual transformation in \cite{chen_simple_2020}. The magnitude for each transformation was stochastically selected from its magnitude range defined in \S\ref{a:transforms} at each iteration. Each of these 16 transformations were evaluated after a short training cycle of $100$ epochs. Each of these transformations were applied on top of a random left-right horizontal flip applied with $p=0.5$.
    \item We further included the five SelfAugment policies from each of its loss functions, evaluated at $100$ epochs.
    \item To further compare with state-of-the-art models, we compare with the MoCoV2 augmentation policy evaluated at ${100, 200}$ epochs 
\end{itemize}
In total, this yields $43$ different models for ImageNet.

\begin{figure*}
\centering
\begin{subfigure}{0.33\textwidth}
    \centering 
    \includegraphics[scale=0.6]{figures/inet_pascal_rotnet.pdf}
\end{subfigure}%
\begin{subfigure}{.33\textwidth}
    \centering
    \includegraphics[scale=0.6]{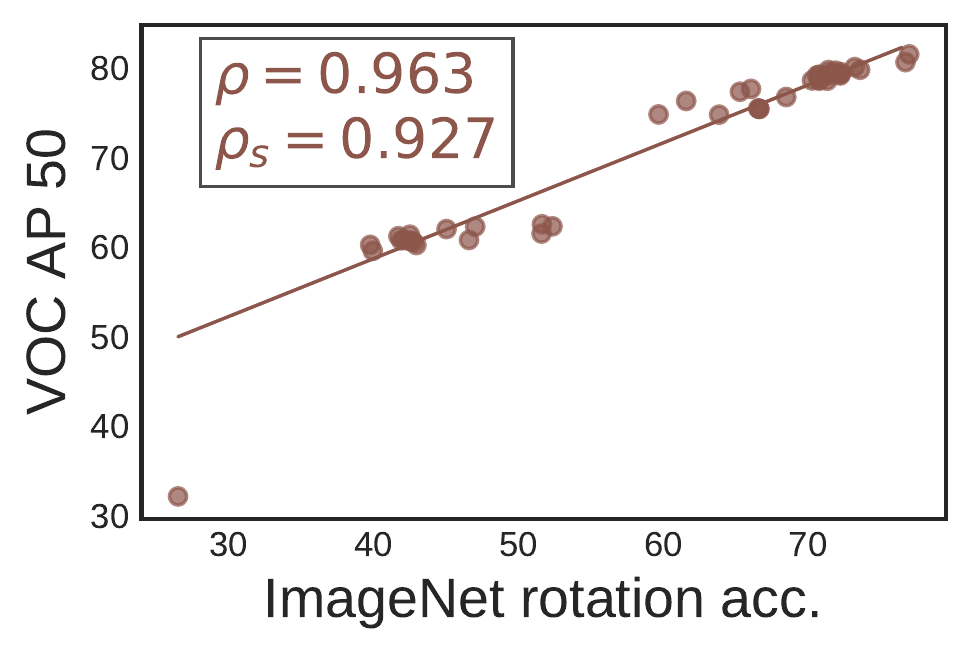}
\end{subfigure}
\begin{subfigure}{.33\textwidth}
    \centering
    \includegraphics[scale=0.6]{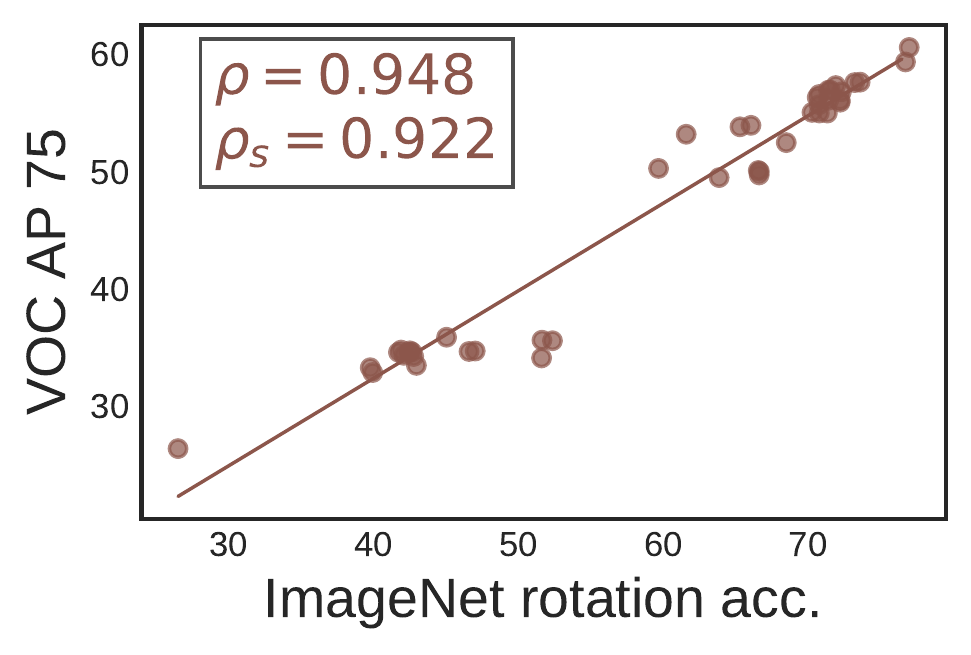}
\end{subfigure}
 
\caption[short]{The ImageNet rotation prediction correlations with transfer performance to VOC07 for all AP, AP$_{50}$, and AP$_{75}$ evaluations. The Spearman rank correlation with rotation prediction is indicated with $\rho$, while the rank correlation with the supervised ImageNet classification performance is indicated with $\rho_s$.}
\label{fig:voc07-detailed}
\end{figure*}

\begin{figure*}[htb]
\centering
\begin{subfigure}{0.33\textwidth}
\centering
\hspace{1em}\textbf{k=1}
\end{subfigure}%
\begin{subfigure}{.33\textwidth}
\centering
\hspace{1em}\textbf{k=4}
\end{subfigure}
\begin{subfigure}{.33\textwidth} 
\centering
\hspace{1em}\textbf{k=8}
\end{subfigure}

\begin{subfigure}{0.33\textwidth}
    \centering
    \includegraphics[width=\textwidth]{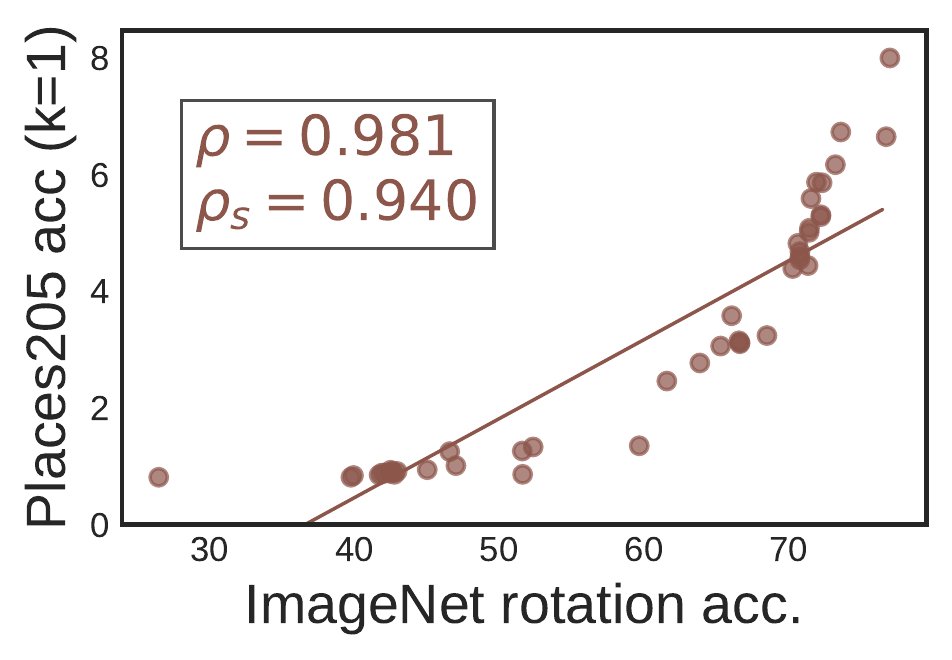}
\end{subfigure}%
\begin{subfigure}{.33\textwidth}
    \centering
    \includegraphics[width=\textwidth]{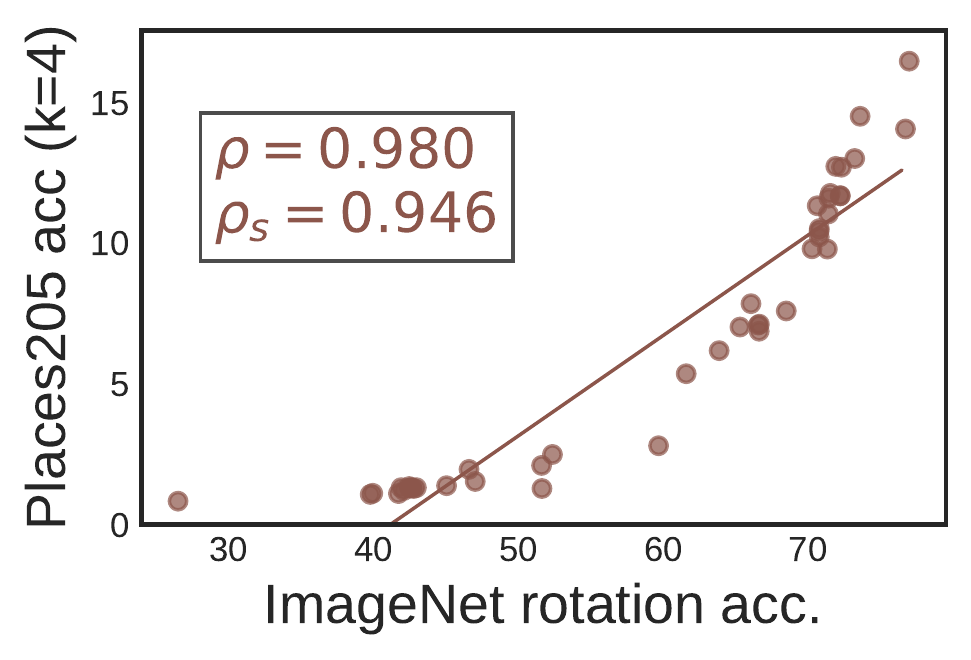}
\end{subfigure}
\begin{subfigure}{.33\textwidth}
    \centering
    \includegraphics[width=\textwidth]{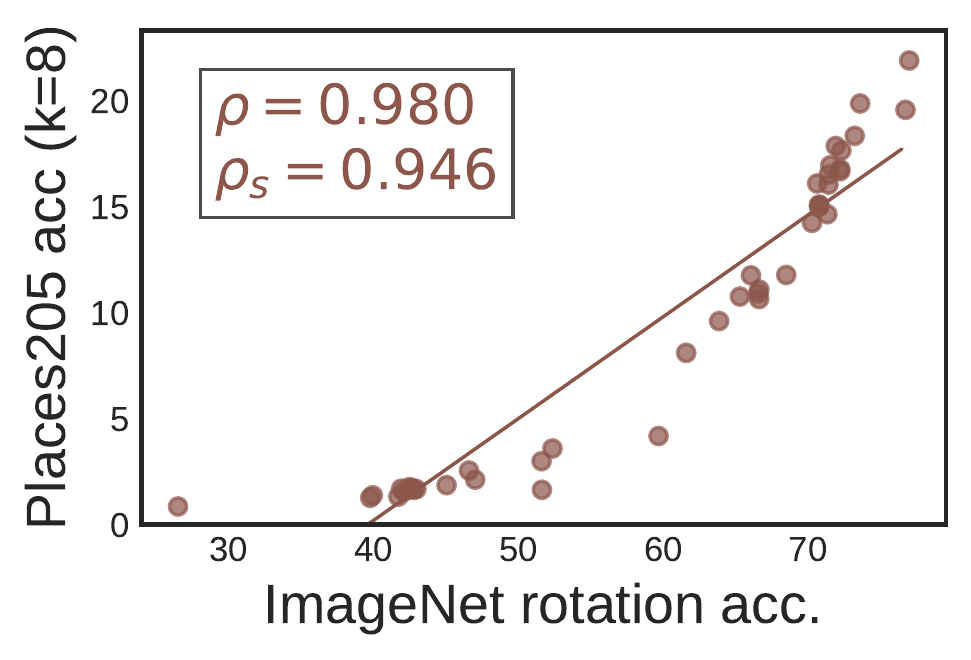} 
\end{subfigure}

\begin{subfigure}{0.33\textwidth} 
\centering
\hspace{1em}\textbf{k=16}
\end{subfigure}%
\begin{subfigure}{.33\textwidth} 
\centering
\hspace{1em}\textbf{k=32}
\end{subfigure}
\begin{subfigure}{.33\textwidth}
\centering
\hspace{1em}\textbf{k=64}
\end{subfigure}

\begin{subfigure}{0.33\textwidth} 
    \centering
    \includegraphics[width=\textwidth]{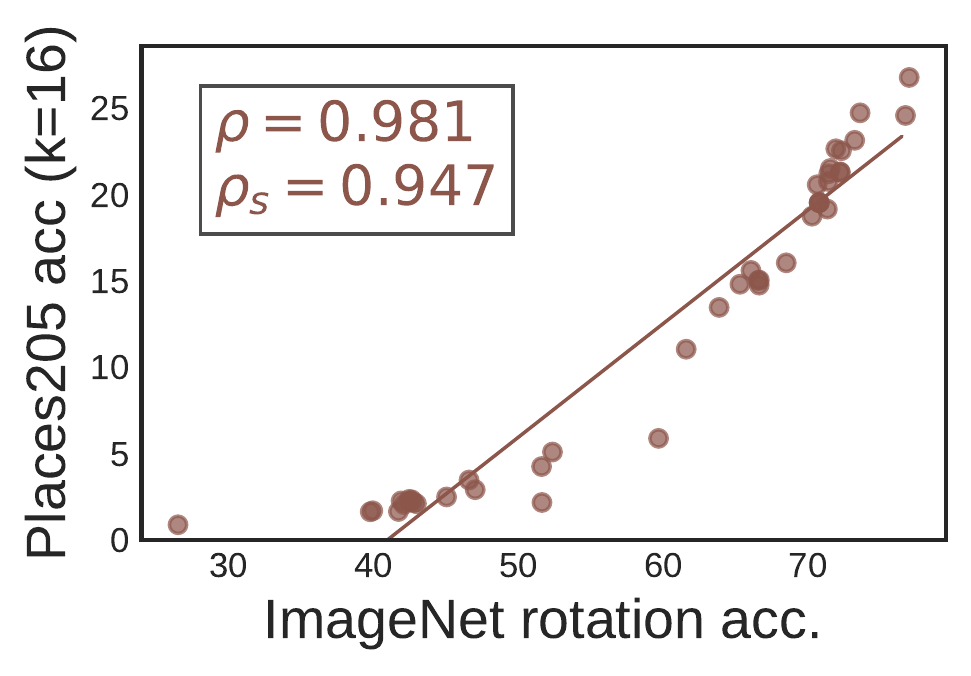}
\end{subfigure}%
\begin{subfigure}{.33\textwidth}
    \centering
    \includegraphics[width=\textwidth]{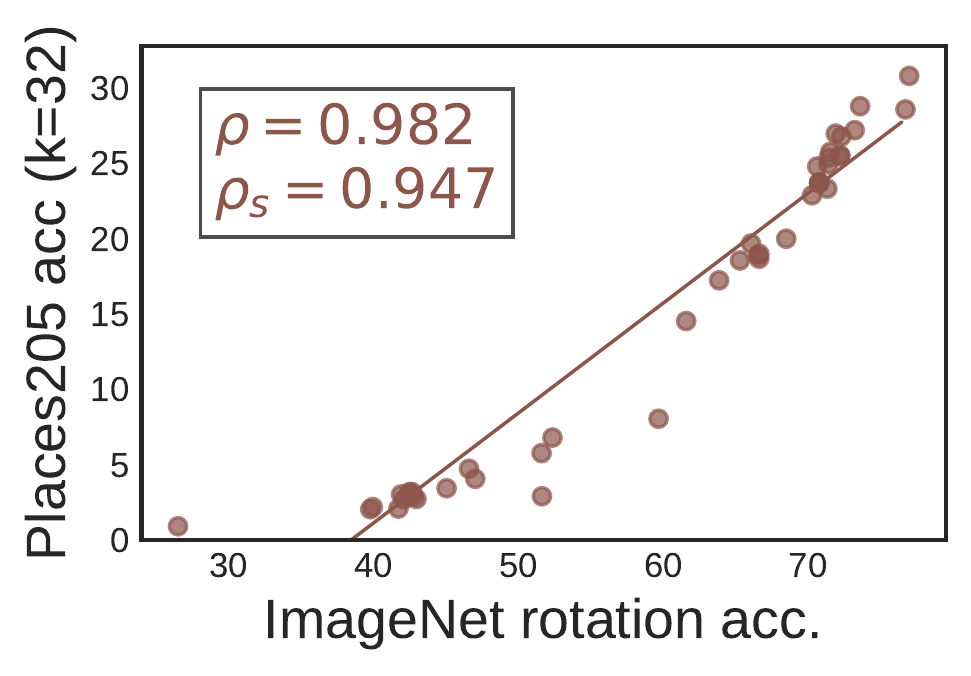}
\end{subfigure}
\begin{subfigure}{.33\textwidth}
    \centering 
    \includegraphics[width=\textwidth]{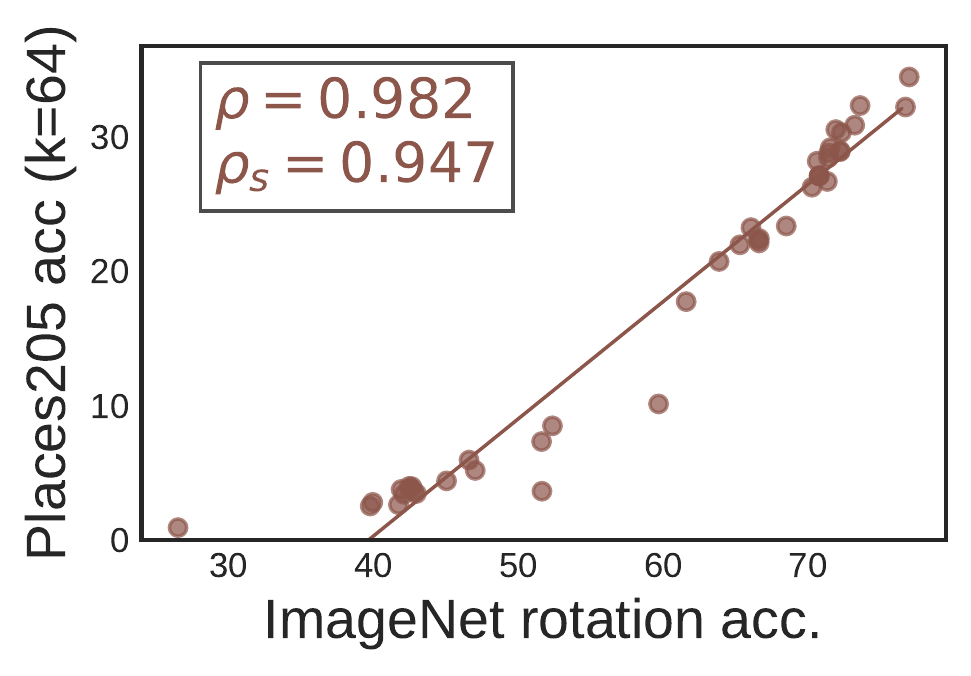}
\end{subfigure}

\caption[short]{The ImageNet rotation prediction correlations with transfer performance to Places205 few label scene classification using $k=\{1,4,8,16,32,64\}$ labels per scene class. The Spearman rank correlation with rotation prediction is indicated with $\rho$, while the rank correlation with the supervised ImageNet classification performance is indicated with $\rho_s$.} 
\label{fig:scenekcorr}
\end{figure*}

Supplementing the main correlation results shown in the experiments section, Figure~\ref{fig:voc07-detailed} shows the the ImageNet rotation prediction correlations with transfer performance to VOC07 for all AP, AP$_{50}$, and AP$_{75}$ evaluations. Figure~\ref{fig:scenekcorr} shows the ImageNet rotation prediction correlations with transfer performance to Places205 few label scene classification using $k=\{1,4,8,16,32,64\}$ labels per scene class. For both VOC07 and Places205, the Spearman rank correlation with rotation prediction is indicated with $\rho$, while the rank correlation with the supervised ImageNet classification performance is indicated with $\rho_s$. Across all evaluation metrics, the rotation correlation is higher than the supervised correlation.

\subsection{Correlation with an MLP rotation prediction}
\label{a:detailcorrmlp}
When following the same evaluation protocol, except using a 2 layer MLP instead of a linear layer for rotation prediction, we find that the CIFAR-10 Spearman rank correlation with the supervised linear classification drops from $0.966$ to $0.904$ while the rotation prediction performance increases by $3.4 \pm 1.4\%$ across all evaluations. This correlation drop indicates that using a simple linear layer, rather than a more complicated network, is ideal for evaluation of the learned representations. A more complicated network can learn its own representation, which distances the evaluation from the learned representations.

\subsection{Correlation demonstration: finding and tuning transformation parameters}

\begin{figure*}[t]
\centering
\begin{subfigure}{0.28\textwidth}
    \centering 
    \includegraphics[scale=0.5]{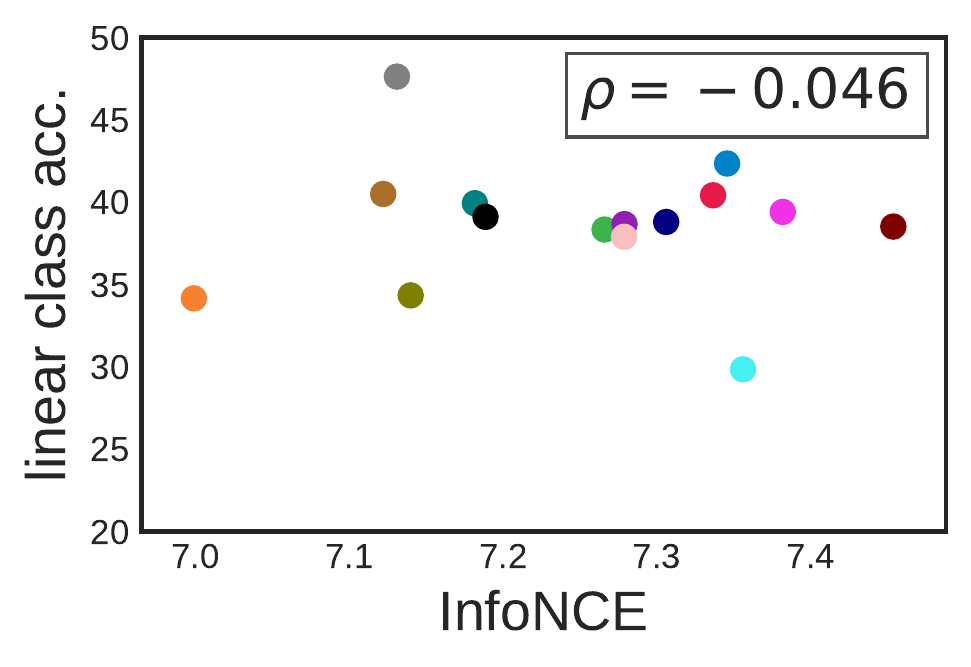}
\end{subfigure}%
\begin{subfigure}{.28\textwidth}
    \centering
    \includegraphics[scale=0.52]{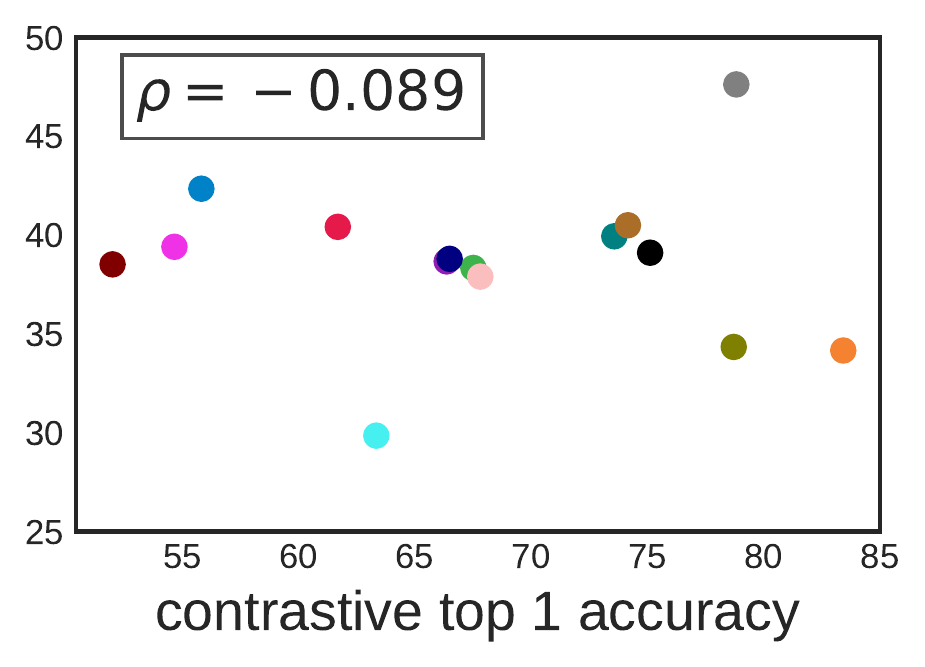}
\end{subfigure}
\begin{subfigure}{.28\textwidth}
    \centering
    \includegraphics[scale=0.52]{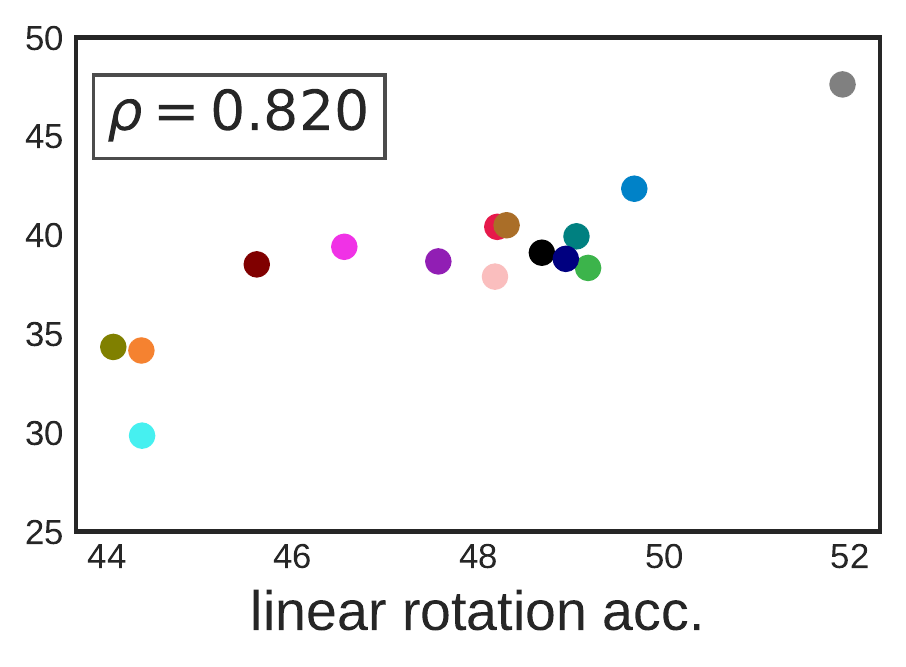}
\end{subfigure}
\begin{subfigure}{.1\textwidth}
    \centering
    \includegraphics[scale=0.4]{figures/aug-legend.pdf}
\end{subfigure}
 
\caption[short]{Similar to Figure~\ref{fig:single-aug}, for CIFAR-10, we plot the supervised classification accuracy (y-axis) vs the InfoNCE loss function (left), contrastive top-1 accuracy (middle), and self-supervised linear rotation accuracy (right), for a self-supervised model trained using one of each transformation used by SelfAugment. Neither of the left two training metrics are a consistent measure of the quality of the representations, while the rotation prediction accuracy provides a strong linear relationship.}
\label{fig:single-aug-cifar-10}
\end{figure*}

Figure~\ref{fig:single-aug-cifar-10} shows the performance of single-transform policies for CIFAR-10. The left and middle plots show the supervised classification accuracy compared with the InfoNCE loss and top-1 contrastive accuracy (how well the instance contrastive model predicts the augmented image pairs), while the right plot shows the rotation prediction accuracy for the image transformations in $\mathbb{O}$ evaluated after $100$ training epochs. Using high or low values of InfoNCE or contrastive accuracy to select the best transformations would select a mixture of mediocre transformations, missing the top performing transformation in the middle. 
By using the rotation prediction, each transformation has a clear linear relationship with the supervised performance, enabling the unsupervised selection of the best transformations.

\begin{figure*}[thb] 
    \centering
        \includegraphics[width=\textwidth]{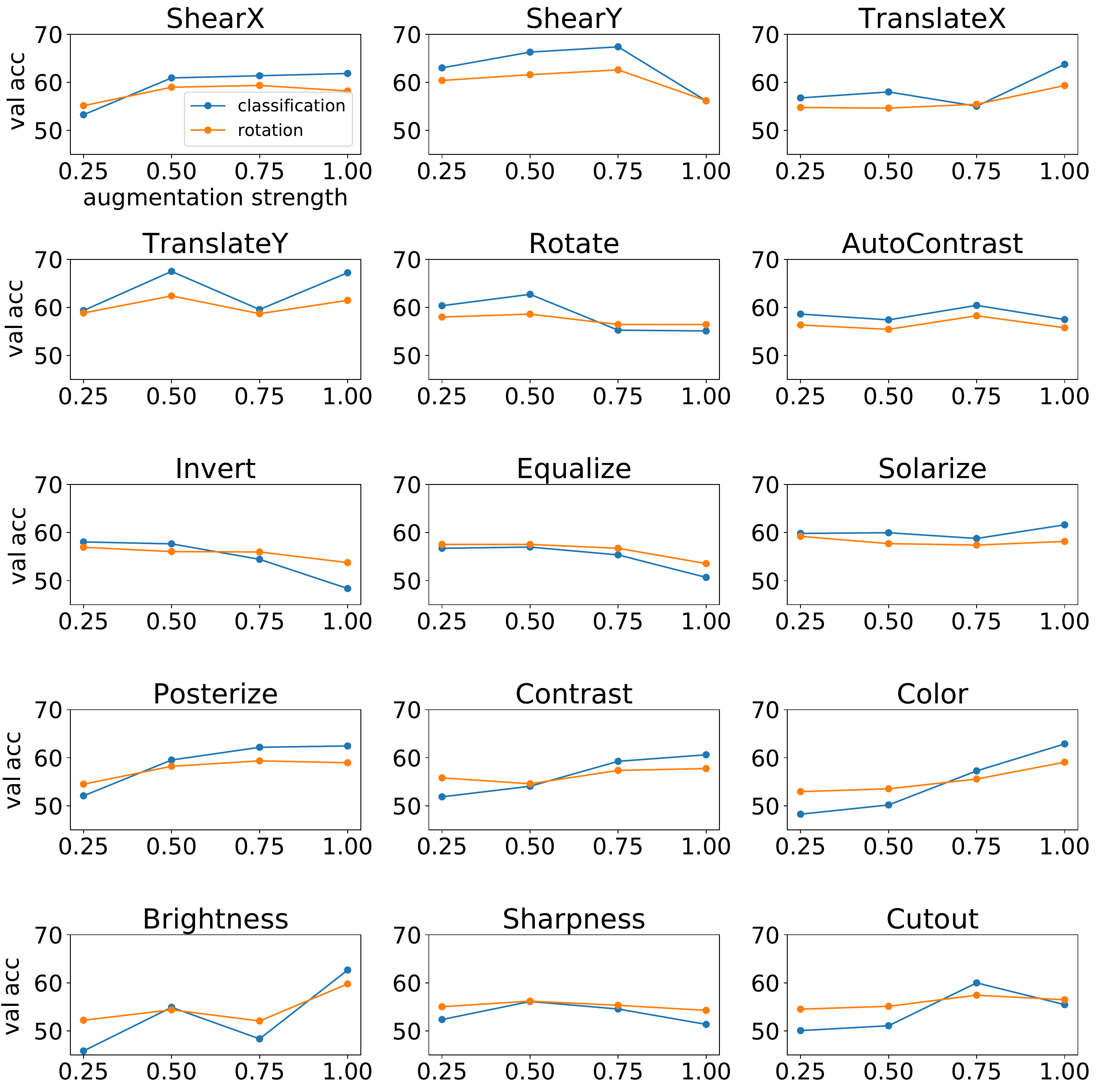}
    \caption[short]{Result of magnitude sweeps on CIFAR-10 for the 15 transformations optimized in SelfAugment. For each augmentation, we train MoCoV2 for 250 epochs using the usual MoCoV2 augmentations, and then added the single augmentation on top. We then vary the range of possible augmentation strength $\lambda$ parameter to be between 0 and the value on the x-axis, and randomly select a value in that range, and apply the augmentation with 100 $\%$ probability. Rotation accuracy and classification accuracy have Spearman rank correlation $\rho = .929$, demonstrating how the relationship between rotation accuracy and classification accuracy can be used to select the parameters of individual transformations.\\\\\\}
    \label{fig:single-aug-corr}
\end{figure*}

Next, Figure~\ref{fig:single-aug-corr} demonstrates that the individual image transformation \emph{parameters} can also be determined through rotation prediction. Specifically, we pre-trained CIFAR-10 using the default MoCoV2 augmentation policy. Then, for each of the fifteen transformation in $\mathbb{O}$, we applied the transformation with probability $1.0$ and a magnitude $\lambda$ randomly selected between 0 and $\{0.25, 0.5, 0.75, 1.0\}$. We evaluate the individual transformation's magnitude parameter using the supervised top-1 linear accuracy (classification) and self-supervised top-1 rotation prediction (rotation) as shown in Figure~\ref{fig:single-aug-corr}. Overall, the supervised linear classification and self-supervised rotation prediction select the the same magnitude parameter for 13 of the 15 transformations and have a strong Spearman rank correlation of $0.929$. Taken together, these results indicate that rotation prediction can be used to both select individual transformations as well as the parameters of the transformations for an augmentation policy.

\subsection{Rotation correlation shows the benefit of Gaussian blur on ImageNet}
In \cite{chen_simple_2020}, the authors conducted a thorough, supervised investigation of a diverse set of image transformations that could be used in an augmentation policy for instance contrastive learning with ImageNet. The \texttt{Gaussian blur} augmentation was shown to be one of the most effective image transformations for ImageNet, and indeed, \cite{chen_improved_2020} released a follow-up paper showing that the MoCo framework~\cite{he2019momentum} significantly benefits from its use. We find that our rotation-based evaluation similarly indicates that Gaussian blur is an effective image transformation for ImageNet under the same conditions studied in \cite{chen_improved_2020}:
\begin{table}[h!]
\small
  \label{table:gblur}
  \centering
  \begin{tabular}{rccc}
Augmentation Policy & Top-1 Supervised Acc. & Rotate Acc. \\
\hline
MoCoV1 & 60.6  & 72.1 \\
MoCoV2 no G-Blur & 63.6 & 74.1 \\
MoCoV2 & 67.7 & 77.0 \\
  \end{tabular}
\end{table}

\label{a:rvsim}

\subsection{Rotation invariant and black-and-white images}
We note that certain types of images are not amenable to certain self-supervised evaluations. 
For instance, rotation evaluation will not work for rotation invariant images (such as images of textures) as the self-supervised evaluation task will not be able to discern the rotations. 
Similarly, a jigsaw task will not be able to discern images with a interchangeable quadrants (such as centered images of flowers), and a color prediction tasks will not work on black and white images (such as x-ray images). 
Therefore, for each image dataset, we recommend using a self-supervised evaluation that does not evaluate an invariant factor of the image dataset, e.g. use rotation prediction if the images are black-and-white. We leave an investigation of the trade-offs between image invariances and self-supervised evaluations to future work.

\clearpage 
\clearpage

\section{Modified RV similarity analysis}
To obtain a better understanding of why the the rotation evaluation has the best correlation with the supervised evaluation performance, we conduct a similarity analysis of the \emph{activations} from the linear evaluation layers. Specifically, we use a modified RV coefficient (as in \cite{similarity}) to measure the similarity between the activations from the rotation, jigsaw, and colorization evaluation layers on top of the frozen encoder network. The RV coefficient is a matrix correlation method that compares paired comparisons $X$ and $Y$ with different number of columns, and is defined as:

\begin{equation}\label{eq:rv_coeff}
     RV(X, Y) = \frac{tr(XX'YY')}{\sqrt{
tr[(XX')^2]tr[(YY')^2]}}
\end{equation}

The RV coefficient approaches 1 when datasets are small, even for random and unrelated matricies. To fix this issue, the modified RV coefficient ($RV_2$) ignores the diagonal elements of $XX'$ and $YY'$, which pushes the numerator to zero when $X$ and $Y$ are random matricies. Hence, the $RV_2$ similarity metric is less sensitive to dataset size. 

\begin{equation}\label{eq:rv_2}\small
     RV_2(X, Y) = \frac{Vec(\Tilde{XX'})'Vec(\tilde{YY'})}{\sqrt{Vec(\Tilde{XX'})'Vec(\Tilde{XX'})  \times Vec(\tilde{YY'}
     )'Vec(\tilde{YY'})}}
\end{equation}

Where $\Tilde{XX'} = XX'-diag(XX')$ and similarly for $\tilde{YY'}$. This metric is invariant to orthogonal transformations and isomorphic scaling, but critically not invariant to arbitrary linear invertible linear transformations between representations (e.g. batch normalization). In \cite{similarity}, the authors show that the $RV_2$ metric could recover expected similarity patterns in neural networks and that it could be used to suggest hypotheses about intermediate representations in deep neural networks. 

\begin{figure*}
\centering
\begin{subfigure}{0.33\textwidth}
    \centering 
    \includegraphics[scale=0.4]{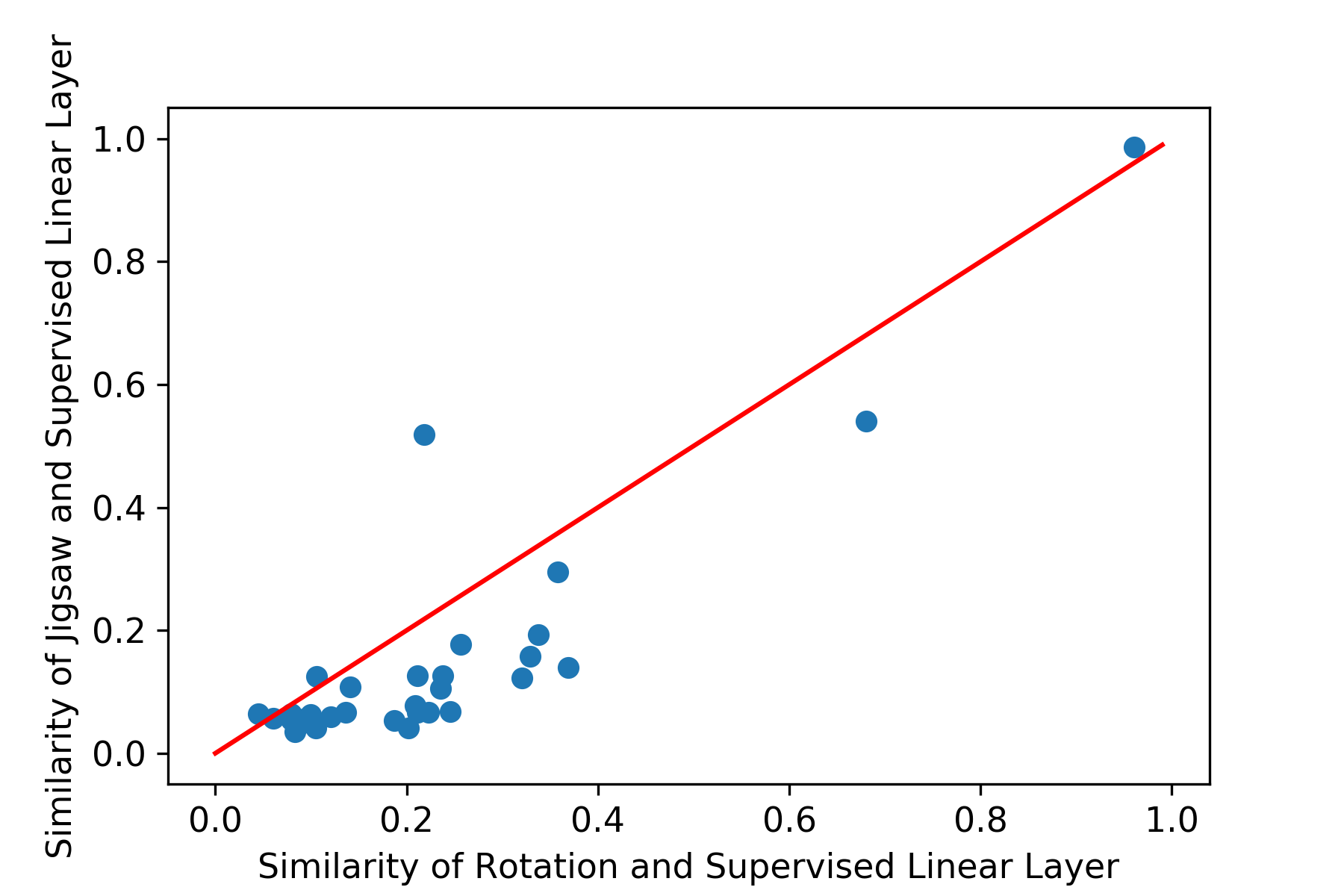}
\end{subfigure}%
\begin{subfigure}{.33\textwidth}
    \centering
    \includegraphics[scale=0.4]{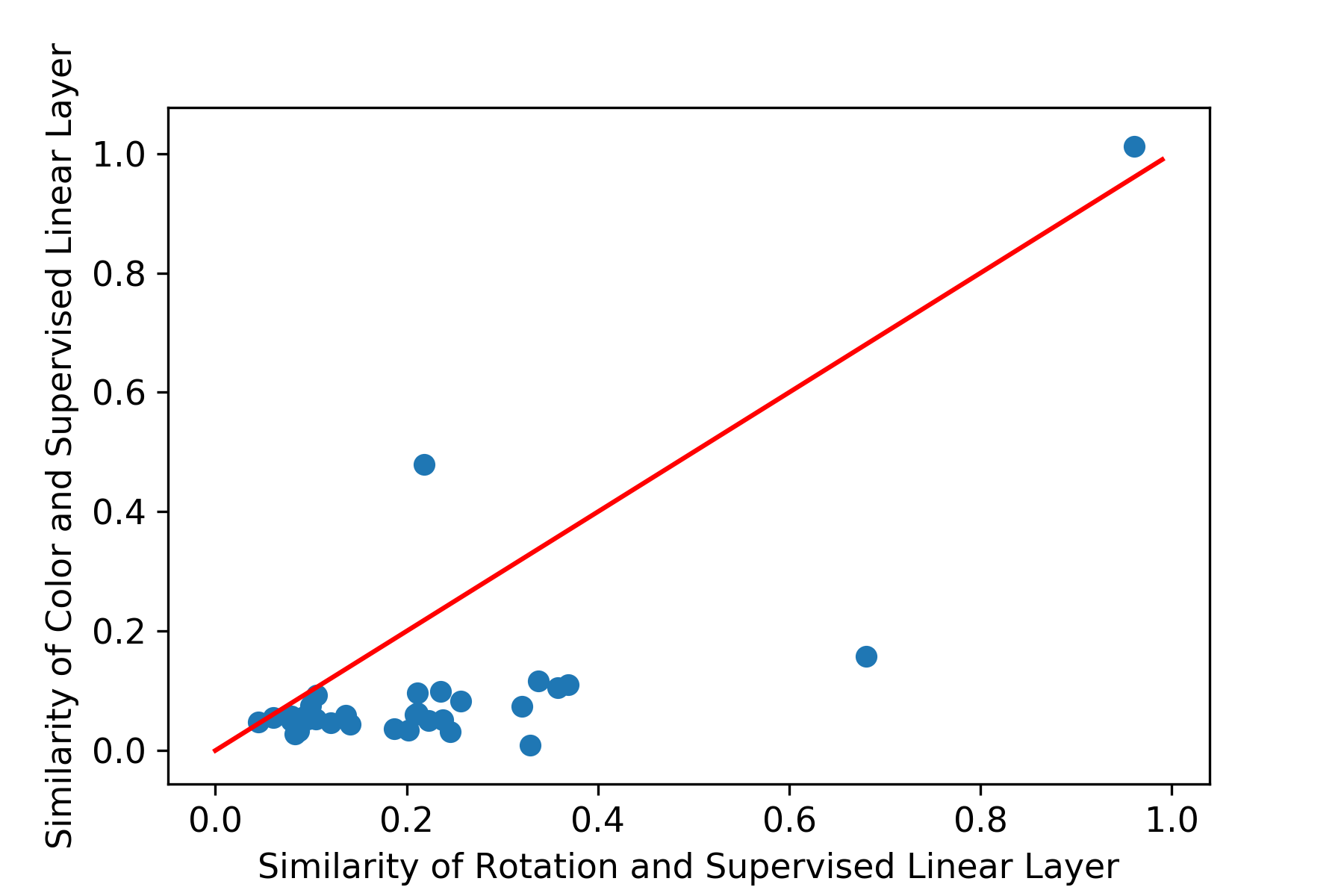}
\end{subfigure}
\begin{subfigure}{.33\textwidth}
    \centering
    \includegraphics[scale=0.4]{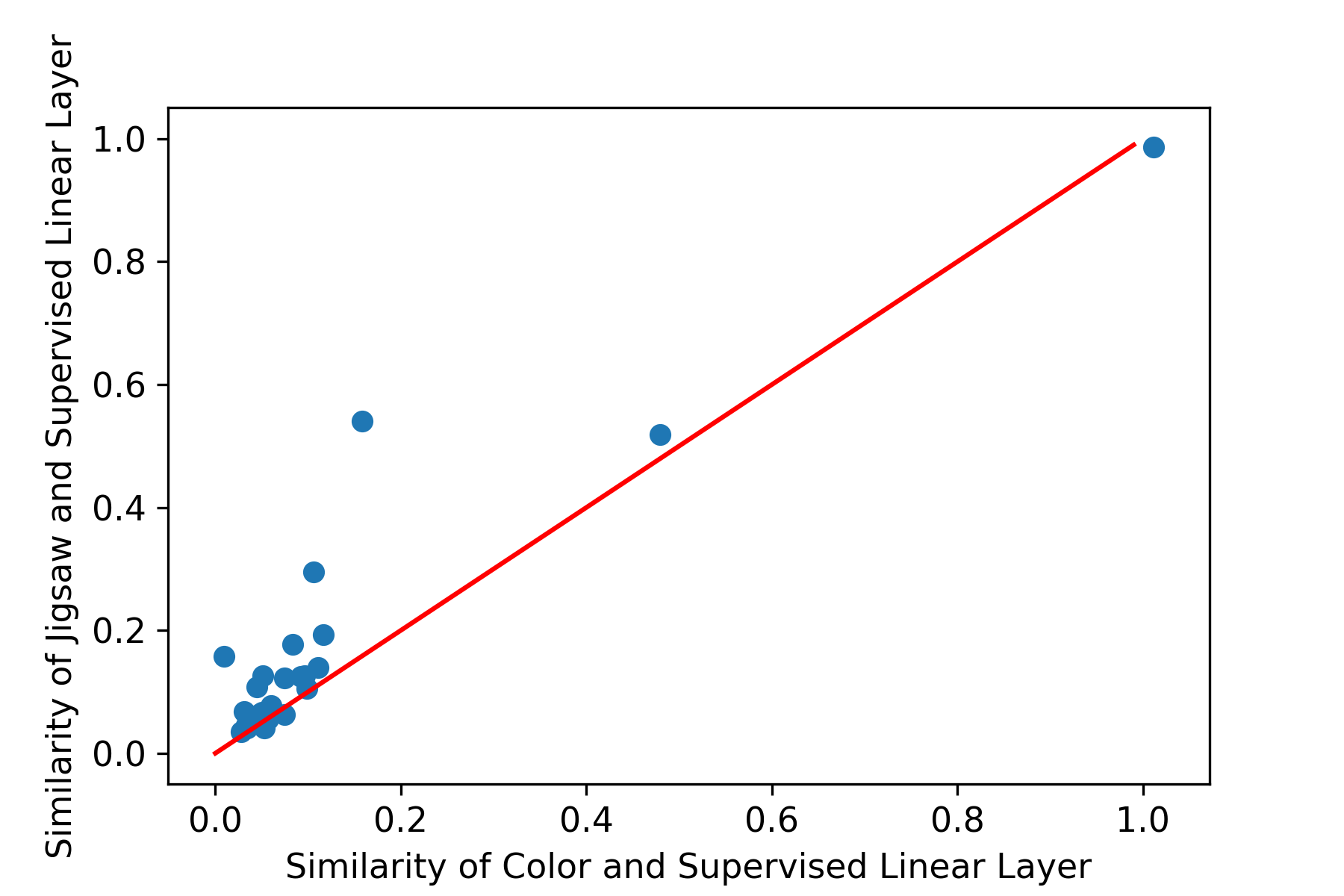}
\end{subfigure}
 
\caption[short]{Comparisons of $RV_2$ Similarity \cite{similarity} between each self supervised task studied and supervised linear layers. Rotation prediction layers have stronger similarity with supervised linear layers than either of the jigsaw and colorization self supervised tasks (p $<$ 2.3x$10^{-5}$, Wilcoxon one-tailed signed rank test). The datapoint with highest similarity was using the strongest settings of SelfRandAugment, which learned poor representations with near chance performance on the supervised downstream task, explaining why similarity was so high across tasks. The strong similarity between rotation prediction and supervised linear layers suggests that they use the underlying representations in a similar way for evaluation, explaining why rotation prediction makes for a strong evaluation metric that correlates more strongly with supervised evaluation than the jigsaw and colorization evaluations.}
\label{fig:rvsim}
\end{figure*}

To study the similarity between linear layers trained using the rotation prediction, jigsaw, and colorization tasks and linear layers trained using supervised learning, we evaluated the $RV_2$ coefficient of 32 different linear layers trained on top of frozen encoders using the CIFAR-10 dataset. Each of the 32 encoders used a different augmentation policy during training. We evaluated the activations at the final linear layer across the entire validation set of CIFAR-10. We used the same set of image transformations before feeding each image into the the network (center crop to 28x28, and normalization across each channel by it's mean and std deviation across the dataset) across all self supervised tasks.  As demonstrated in Figure \ref{fig:rvsim}, activations from rotation prediction layers had significantly stronger similarities with activations from supervised layers than other self supervised tasks. This demonstrates that the rotation prediction task uses the learned representation in a significantly more similar way for evaluation compared to the other evaluation tasks, and provides evidence that rotation evaluation performance not only correlates, but so does the activations from the evaluation layer.

\section{Augmentation policy exploration via Bayesian optimization}
\label{a:bayesopt}
As in \cite{lim_fast_2019}, we used policy exploration search to automate the augmentation search. Since there are an infinite number of possible policies, we applied Bayesian optimization to explore augmentation strategies. In line 11 in Algorithm 1, we employed the Expected Improvement (EI) criterion as an acquisition function to explore $\mathcal{B}$ candidate policies efficiently: $EI(\mathcal{T}) = \mathbb{E}[min(\mathcal{L(\theta, \phi | \mathcal{T}(\mathcal{D}_{\mathcal{A}})} - \mathcal{L^{\dagger}}, 0)]$. Here $L^{\dagger}$ represents a constant threshold determined by the quantile of observations amongst previously explored policies. As in \cite{lim_fast_2019}, we used variable kernel density estimation on a graph-structured search space to approximate the criterion. Since this method is already implemented in the tree-structured Parzen estimator algorithm we used \texttt{Ray}\footnote{\url{https://github.com/ray-project/ray}} and \texttt{Hyperopt} to implement this in parallel. 

In \cite{lim_fast_2019}, the authors try to align distributions of data using supervised loss, then retrain the network using supervised loss. Since we do not directly retrain with the loss functions used to find augmentations, our method can instead be thought of as finding distributions of the data - via augmentation policies - that minimize (or maximize) alignment as defined by our loss functions. 

The full list of augmentations and range of magnitudes explored during augmentation policy exploration are detailed in Table \ref{table:aug}.

\section{SelfAugment Loss Functions}
\label{a:lossfun}

\begin{figure*}[t]
\centering
    \includegraphics[scale=0.6]{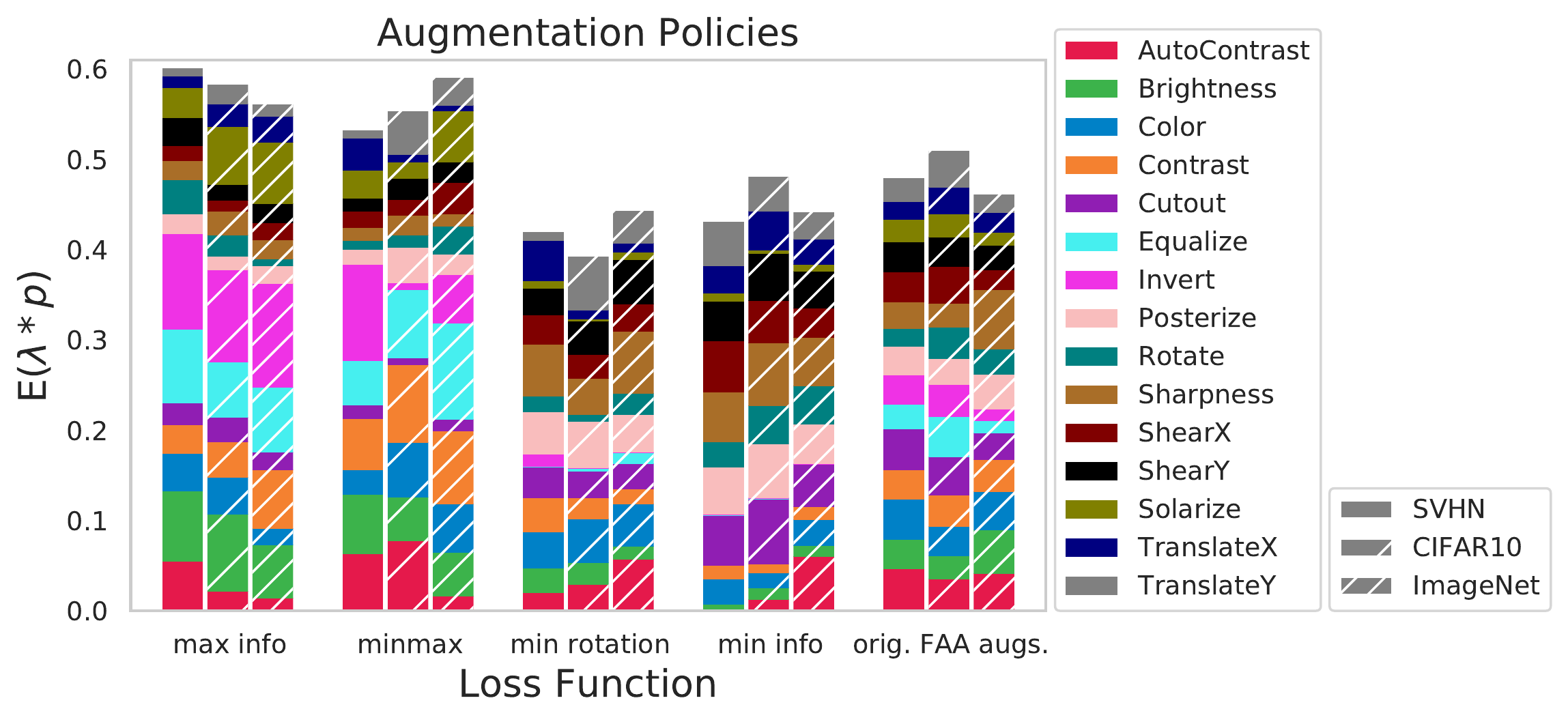}
\caption[short]{Visualization of the augmentation policies found with SelfAugment as well as Fast AutoAugment for supervised learning. This figure shows expected augmentation strength (mean magnitude$\times$normalized probability) of each augmentation, evaluated across all datasets and loss functions. As expected, minimizing InfoNCE results in augmentations with smaller magnitude, and emphasizes augmentations that do not alter the image much (e.g. \texttt{Sharpness}). Maximizing InfoNCE results in augmentations with a larger magnitude and emphasizes augmentations (e.g. \texttt{Invert, Solarize}) which heavily alter the image. The minimax loss function yields an augmentation policy that strikes a middle ground between the two, with strong augmentations, but reduced emphasis on heavy augmentations like \texttt{Invert}. Comparison of the policies that worked best with contrastive learning relative to the original FAA policies reveals that our policies are (i) stronger, and (ii) have more variability in a single augmentation's E($\lambda$*$p$) relative to policies that were used for supervised learning.}
\label{fig:aug-magnitudes}
\end{figure*}

In this section we discuss the loss functions used for SelfAugment in greater detail and their resulting augmentation policies which are summarized in Figure~\ref{fig:aug-magnitudes}.

\begin{itemize}
\item \textbf{Min.~eval error}: $$\mathcal{T}^{SS} =  {\text{argmin}_\mathcal{T}}\mathcal{L}_{\text{SS}}(\theta_{\mathcal{M}},\phi_{\text{ss}}|\mathcal{T}(D_{\mathcal{A}}))$$ where $\mathcal{L}_{\text{SS}}$ is the self-supervised evaluation loss, which yields policies that should result in improved performance of the evaluation if we were to retrain the linear classifier with the selected augmentations. However, since the augmentations are instead used to create a contrastive learning task, it is important that we find a different set of augmentations that take the contrastive task into consideration. Indeed, using this loss function results in weak expected augmentation strengths (see Figure~\ref{fig:aug-magnitudes}), resulting in relatively poor downstream performance (see Table ~\ref{table:results}).

\item \textbf{Min.~InfoNCE}: $$\mathcal{T}^{\text{I-min}} =  {\text{argmin}_\mathcal{T}}\mathcal{L}_{\text{NCE}}(\theta_{\mathcal{M}}|\mathcal{T}(D_{\mathcal{A}}))$$ where $\mathcal{L}_{\text{NCE}}$ is the InfoNCE loss from Eq.~\ref{eq:infonce_loss}, which yields policies that make it easier to distinguish image pairs in the contrastive feature space. This loss function should results in small magnitude augmentations, since a trivial way to minimize InfoNCE is to apply no transforms - resulting in trivial minimization of the InfoNCE loss. Indeed, the loss function yields (i) lower  expected augmentation strengths and (ii) emphasizes transformations like \texttt{Sharpness}, Figure ~\ref{fig:aug-magnitudes}. We did not expect this loss function to perform well, and was primarily included as a sanity check that minimizing InfoNCE would result in light augmentations.

\item \textbf{Max InfoNCE}: $$\mathcal{T}^{\text{I-max}} =  {\text{argmin}_\mathcal{T}}-\mathcal{L}_{\text{NCE}}(\theta_{\mathcal{M}}|\mathcal{T}(D_{\mathcal{A}}))$$ negates the previous loss function, yielding policies that make it difficult to distinguish image pairs in the feature space. In practice, this results in high magnitude augmentations and emphasizes augmentations like \texttt{Invert}. We hypothesized that maximizing InfoNCE could result in strong augmentations that could create a challenging contrastive task. However, the augmentations do not have any regularizing that would ensure the image maintains important features, leading to relatively suboptimal performance (see Table~\ref{table:results}).

\item \textbf{Min} $\mathcal{L}_{\text{ss}}$ \textbf{max} $\mathcal{L_{\text{NCE}}}$: $\mathcal{T}^{\text{minmax}} =  {\text{argmin}_\mathcal{T}}\mathcal{L}_{\text{ss}} - \mathcal{L}_{\text{NCE}}$ 
yields policies with difficult transformations that maximize InfoNCE, while maintaining salient object features that minimize the self-supervised evaluation. When using a linear rotation evaluation prediction, we found this loss function to have the strongest performance for contrastive learning ($\S\ref{experiments}$). In practice, we normalized $\mathcal{L}_{\text{rot}}$ and  $\mathcal{L_{\text{NCE}}}$ by their expected value for the training data across the K-folds when training with the base augmentation, since $\mathcal{L_{\text{NCE}}}$ was usually higher than $\mathcal{L}_{\text{rot}}$, but had each loss contribute equally for augmentation policy selection. Policies that optimized this loss emphasized transformations like \texttt{Equalize, AutoContrast} and \texttt{Contrast} more than others. For ImageNet, these transformations proved to be challenging yet useful. Notably, they closely resemble the effects of MoCov2 and SimCLR's Color Jitter \cite{chen_improved_2020}. 
\end{itemize}

Finally, while \textbf{min} $\mathcal{L}_{\text{rot}}$ \textbf{max} $\mathcal{L_{\text{NCE}}}$ works well, one could potentially gain performance by weighting the importance of minimizing $\mathcal{L}_{\text{rot}}$ vs maximizing $\mathcal{L_{\text{NCE}}}$. Hence we propose experimenting with different values of $\lambda_{\text{NCE}}$ and $\lambda_{\text{rot}}$ when optimizing the following objective: \textbf{min} $\lambda_{\text{rot}}\mathcal{L}_{\text{rot}}$ \textbf{max} $\lambda_{\text{NCE}}\mathcal{L_{\text{NCE}}}$.

\section{Using the Original MoCoV2 \cite{chen_improved_2020} policy as the base policy}
We replicated our SelfAugment results in Table \ref{table:results} for the CIFAR10 dataset, where instead of using a single augmentation as the base policy, we used the full augmentation policy from \cite{chen_improved_2020} as the base policy. Results are shown in Table \ref{table:ontopofmoco}. This improved performance when we used the augmentation policies learned using min  $\mathcal{L}_{\text{rot}}$ as feedback. Minimizing rotation may be the best approach for learning augmentation policies on top of established augmentation policies, because it produces augmentations that would improve performance if we retrained the linear classifier with the selected augmentations. Since the augmentation policy for contrastive learning is already quite strong when we use the MoCov2 augmentations as the base policy, this likely allows us to focus on improving generalization \cite{lim_fast_2019}, leading to improved performance. Interestingly, the other SelfAugment Loss functions (see Appendix \ref{a:lossfun} for more detail) all deteriorated performance. This is likely because trying to change the contrastive performance of such a finely tuned set of augmentations requires more careful tuning. It is possible the minimax approach may result in improved performance with more careful tuning of the weighting of the rotation and contrastive losses. 

It is also worth noting that using the best policy using the MoCov2 augmentations as the base augmentation set, then using SelfAugment with  min  $\mathcal{L}_{\text{rot}}$ as feedback slightly outperformed the full SelfAugment pipeline with  min  $\mathcal{L}_{\text{rot}}$ max $\mathcal{L}_{\text{ICL}}$ as feedback. For datasets where researchers are confident that the augmentations from \cite{chen_improved_2020} are a strong base set of augmentations, this approach is worth exploring, and can be easily done using our pipeline. However, it is not possible when a good base augmentation policy is unknown. We leave further exploration of this approach as future research, as we chose to focus on the case where no existing augmentation policy is known.

\begin{table}[h!]

\small

  \centering
  \begin{tabular}{rc}
    \hline
    \textbf{unsup. feedback} & C10 \\
    \hline
    SelfAug (min rot) & \textbf{92.8} \\
    SelfAug (min Info) & 92.2 \\
    SelfAug (max Info) & 90.5 \\
    SelfAug (minimax) & 90.9 \\
    \hline
    \textbf{supervised feedback} \\
    \hline
    MoCov2\cite{chen_improved_2020} & 92.3
  \end{tabular}
  \caption{Results on CIFAR10 when using SelfAugment but using the augmentation policy from \cite{chen_improved_2020} as the base policy. Because the base augmentation policy is already strong, minimizing rotation loss during the augmentation policy search produces the best results. The results with the MoCoV2 augmentations as the base policy and minimization of rotation prediction slightly outperform the best results using SelfAugment and the minimax loss as feedback (92.6, see Table \ref{table:results}).}
  \label{table:ontopofmoco}
\end{table}



\section{Computational Efficiency}
\label{a:eff}
 
SelfAugment can be broken up into three steps:
\begin{enumerate}
    \item \textbf{Finding the base augmentation}. This entails training 16 instances of MoCoV2 for 10-15\% of the total epochs typically used for pre-training, using all of the training data available.  
    \item \textbf{Training on K-Folds using the base augmentation}. For CIFAR-10/SVHN we used the entire training dataset for this step, and evaluated for the same number of epochs as we use to train the final models, using all the training data from each fold. For ImageNet, we used a subset of 50,000 images and trained for 5x longer, to make up for the smaller amount of images. This was done to improve the computational efficiency of step $3$.
    \item \textbf{Finding augmentation policies}. Because we only need to complete forward passes of the network, this is relatively efficient. We use the held out data from each of the K-folds to evaluate the model with different augmentation policies applied. 
\end{enumerate}

It is important to note that steps 1 and 2 are \emph{shared} across all augmentation policies found with SelfAug, meaning that they do not need to be repeated to find a new augmentation policy using a different loss function. This could allow for rapid experimentation with new loss functions for SelfAugment in the future. 

Meanwhile, SelfRandAug's computational time is completely determined by the user-defined search space over $\lambda$ and $N_{\tau}$. However, because it requires training multiple models to completion on the entire dataset, it is more computationally intensive than SelfAug. SelfRandAug's computation time could be reduced by training on a subset of Images in a large training set.

Finally, it is worth noting that evaluating a \textit{single} augmentation's various hyperparameters for MoCoV2 can be extremely computationally expensive. Because pre-training MoCoV2 for 100 epochs then training a linear head for evaluation takes a total of 244 GPU Hours on a Tesla V100 GPU, searching over various augmentations and their strength and probability parameters can easily require thousands of GPU hours. Hence, an additional contribution of this work is providing a fast, automatic method for selecting augmentations for instance contrastive learning.

\begin{table}[htb]
\label{table:eff}
\small
  \caption{Computational time, measured in one hour on one NVIDIA Tesla V100 GPU for SelfAugment and SelfRandAug on ImageNet. SelfAug times are evaluated when using the minimax loss function, which takes the most time because we evaluate both InfoNCE and rotation loss. For SelfRandAug, computational time reflects our grid search over $\lambda=\{5, 7, 9, 11, 13\}$, $N_\tau=2$, plus the time to train a rotation head on top of each representation. This time could be larger or smaller depending on the parameters $\lambda$ and $N_\tau$ a user wants to search over.}
  \label{table:effresults}
  \centering
  \begin{tabular}{lllll}
\hline
\textbf{Algorithm} & Find Base Aug & Train & Find Aug & \textbf{Total} \\
\hline
\textbf{SelfAug} & 476.6 & 155.0 & 97.7 & \textbf{729.4}\\
\textbf{SelfRandAug} & & &  & \textbf{1220.0} \\

  \hline
  \end{tabular}
\end{table}

\end{document}